\documentclass[10pt,twocolumn,letterpaper]{article}
\usepackage{cvpr}              

\usepackage{graphicx}
\usepackage{amsmath}
\usepackage{amssymb}
\usepackage{booktabs}
\usepackage{multirow}

\usepackage[pagebackref,breaklinks,colorlinks]{hyperref}

\usepackage[capitalize]{cleveref}
\crefname{section}{Sec.}{Secs.}
\Crefname{section}{Section}{Sections}
\Crefname{table}{Table}{Tables}
\crefname{table}{Tab.}{Tabs.}

\renewcommand{\eqref}[1]{Eq.~(\ref{#1})}

\def\ie{{i.e.,~}}
\def\eg{{e.g.,~}}

\def\etal{{et al.~}}

\def\cvprPaperID{2006} 
\def\confName{CVPR}
\def\confYear{2022}

\begin{document}

\title{Interactive Image Synthesis with Panoptic Layout Generation}

\renewcommand{\thefootnote}{\fnsymbol{footnote}}
\author{Bo Wang \quad Tao Wu \quad Minfeng Zhu \quad Peng Du$\footnotemark[1]$ \\[4pt]
Huawei Technologies\\[4pt]
{\tt\small \{wangbo341, zhuminfeng, dupeng25\}@hisilicon.com, taowu1@huawei.com}
}
\maketitle
\footnotetext[1]{Corresponding author.}
\begin{abstract}
Interactive image synthesis from user-guided input is a challenging task when users wish to control the scene structure of a generated image with ease. Although remarkable progress has been made on layout-based image synthesis approaches, existing methods require high-precision inputs such as accurately placed bounding boxes, which might be constantly violated in an interactive setting. When placement of bounding boxes is subject to perturbation, layout-based models suffer from ``missing regions'' in the constructed semantic layouts and hence undesirable artifacts in the generated images. In this work, we propose Panoptic Layout Generative Adversarial Network (PLGAN) to address this challenge. The PLGAN employs panoptic theory which distinguishes object categories between ``stuff'' with amorphous boundaries and ``things'' with well-defined shapes, such that stuff and instance layouts are constructed through separate branches and later fused into panoptic layouts. In particular, the stuff layouts can take amorphous shapes and fill up the missing regions left out by the instance layouts. We experimentally compare our PLGAN with state-of-the-art layout-based models on the COCO-Stuff, Visual Genome, and Landscape datasets. The advantages of PLGAN are not only visually demonstrated but quantitatively verified in terms of inception score, Fr\'echet inception distance, classification accuracy score, and coverage.
The code is available at \href{https://github.com/wb-finalking/PLGAN}{https://github.com/wb-finalking/PLGAN}.
\end{abstract}

\section{Introduction}

\begin{figure}[t]
\begin{center}
{\tabcolsep0.01in
\begin{tabular}{cccc}
\rotatebox{90}{\quad \quad Grid2Im} 
& \includegraphics[width=0.301\linewidth]{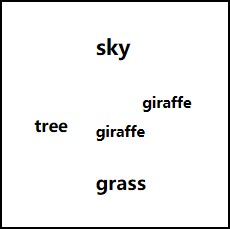}
& \includegraphics[width=0.3\linewidth]{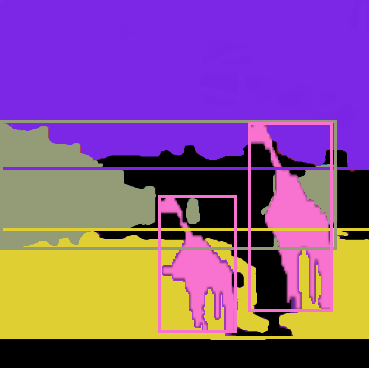} 
& \includegraphics[width=0.3\linewidth]{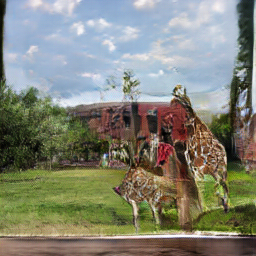}\\
&Object Input &Instance Layout & Generated Image\\
\rotatebox{90}{\quad \quad PLGAN}
& \includegraphics[width=0.3\linewidth]{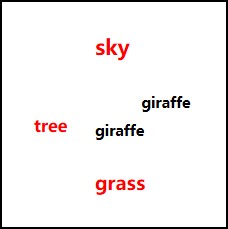} 
& \includegraphics[width=0.302\linewidth]{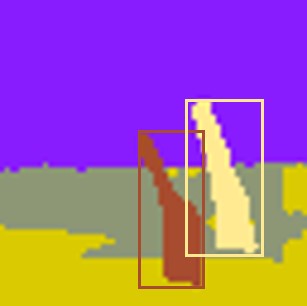} 
& \includegraphics[width=0.302\linewidth]{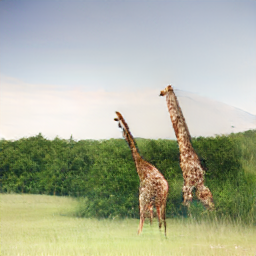} \\
&Instance/Stuff Input &Panoptic Layout & Generated Image\\
\end{tabular}}
\end{center}
\vspace{-.5cm}
\caption{Scene-to-image synthesis by Grid2Im \cite{ashual2019specifying} vs.~PLGAN. 
Unlike the instance layout utilized by Grid2Im that treats all objects as instances (things), 
the panoptic layout by PLGAN distinguishes objects between \textbf{instances} and \textcolor{red}{\textbf{stuff}},
thus eliminating artifacts in the \emph{missing regions} (marked black in the layout).
}
\label{fig:head}
\end{figure}

Tremendous progress has been made on conditional image synthesis for creative design. 
Among different formats of conditional inputs are categories~\cite{mirza2014conditional, Karras2019stylegan2, zhang2019self, brock2018large, miyato2018cgans, odena2017conditional}, source images~\cite{Li2021CVPR, zhu2017unpaired, kim2017learning}, text description~\cite{Xia2021CVPR, reed2016generative, zhang2017stackgan}, scene graphs~\cite{johnson2018image,ashual2019specifying,Yang2021CVPR} and semantic layouts~\cite{park2019semantic, wang2018high, zhu2020sean}. 
To date, \emph{interactive image synthesis} via conditional generative models remains a contemporary challenge.
Text-to-image models usually suffers from reasoning object locations and relations \cite{zhang2017stackgan}. 
Image synthesis from semantic layouts provides an alternative way for computer-human interaction \cite{li2019object, hong2018inferring} and yields aesthetically pleasing results \cite{park2019semantic, wang2018high, zhu2020sean}. However, high-quality semantic layouts demand professional skills in free-hand drawing from scratch by users, which prevent them from being used as a drag-and-drop GUI for novice users.
In this respect, scene graph has attracted much attention in recent years \cite{johnson2018image, ashual2019specifying}, as it requires only multiple object placements on the artboard and allows user-friendly manipulation with individual objects. 

A milestone in scene-graph-to-image synthesis is made by Grid2Im \cite{ashual2019specifying}, which roughly consists of two stages: layout construction and image generation.
First, the input conditions are passed to construct an instance layout with per-object masks and bounding boxes.
Secondly, conditioned on the instance layout a photo-realistic image is sythesized as the outcome of the image generation stage.
Whilst Grid2Im \cite{ashual2019specifying} requires ground truth segmentation maps as supervision signals, LostGAN \cite{sun2019image, sun2020learning} can learn the intermediate instance layout in a weakly-supervised way.

Besides \cite{ashual2019specifying, sun2019image, sun2020learning}, instance layout-based generative models such as \cite{johnson2018image, zhao2019image, li2019object, hong2018inferring} have also driven progress on many cross-domain image synthesis tasks.
A common caveat among aforementioned methods consists in that they are sensitive to spatial perturbation of scene objects and suffer from the \emph{region missing} problem, especially in interactive scenarios.
They may predict intermediate layouts with empty areas where pixels may not have correct category information.
During the training phase, ground-truth bounding boxes and masks usually covers the whole image lattice.
However, in interactive scenarios users can place objects with bounding boxes arbitrarily.
In addition, the predicted per-object masks will not fill up the corresponding bounding boxes. 
Therefore, the intermediate layout may not be covered completely by object masks, yielding the region missing problem.
An illustration of region missing by Grid2Im \cite{ashual2019specifying} is displayed in Fig.~\ref{fig:head}.
Unsurprisingly, an imperfect semantic layout containing empty areas induces undesirable artifacts in the generated image.

In this work, we propose \emph{Panoptic Layout Generative Adversarial Network} (PLGAN) for interactive image synthesis.
Different from prior works that treat all objects invariably as instances (or things), we employ the panoptic segmentation theory \cite{kirillov2019panoptic} that splits objects into uncountable \emph{stuff} (which refers to amorphous background such as grass, sky or sea) and countable \emph{things} (which are foreground objects with well-defined shapes such as people, animals or vehicles).
Furthermore, we develop the panoptic layout generation (PLG) module, which utilizes in parallel a stuff branch for stuff layout construction and an instance branch for instance layout construction. The instance branch predicts per-instance bounding boxes and maskes as in \cite{ashual2019specifying, sun2019image}.
The stuff branch generates pixel-wise masks for all stuff objects that cover the whole image lattice.
Then the instance layout and the stuff layout are combined into a panoptic layout through an Instance- and Stuff-Aware Normalization (ISA-Norm) module.
Images synthesis from such a panoptic layout successfully eliminates missing regions and behaves more robust to perturbation of object locations; see Fig.~\ref{fig:head} for a visualized example.

Our contributions are summarized as follows:
\begin{itemize}
   \item We leverage panoptic layouts in interactive image synthesis to resolve the region missing problem inheritted by current instance layout-based approaches.
   \item Regarding model architecture, we propose to separate treatment of stuff and thing objects during layout constructions and later fuse the constructed instance and stuff layouts into a panoptic layout via Instance- and Stuff-Aware Normalization (ISA-Norm).
   \item Our experiments show qualitative and quantitative comparisons on the COCO-Stuff dataset, Visual Genome, and Landscape datasets, and demonstrate of the merits of our PLGAN over state-of-the-art layout-based approaches.
\end{itemize}

\begin{figure*}[t]
   \begin{center}
   \includegraphics[width=0.85\linewidth]{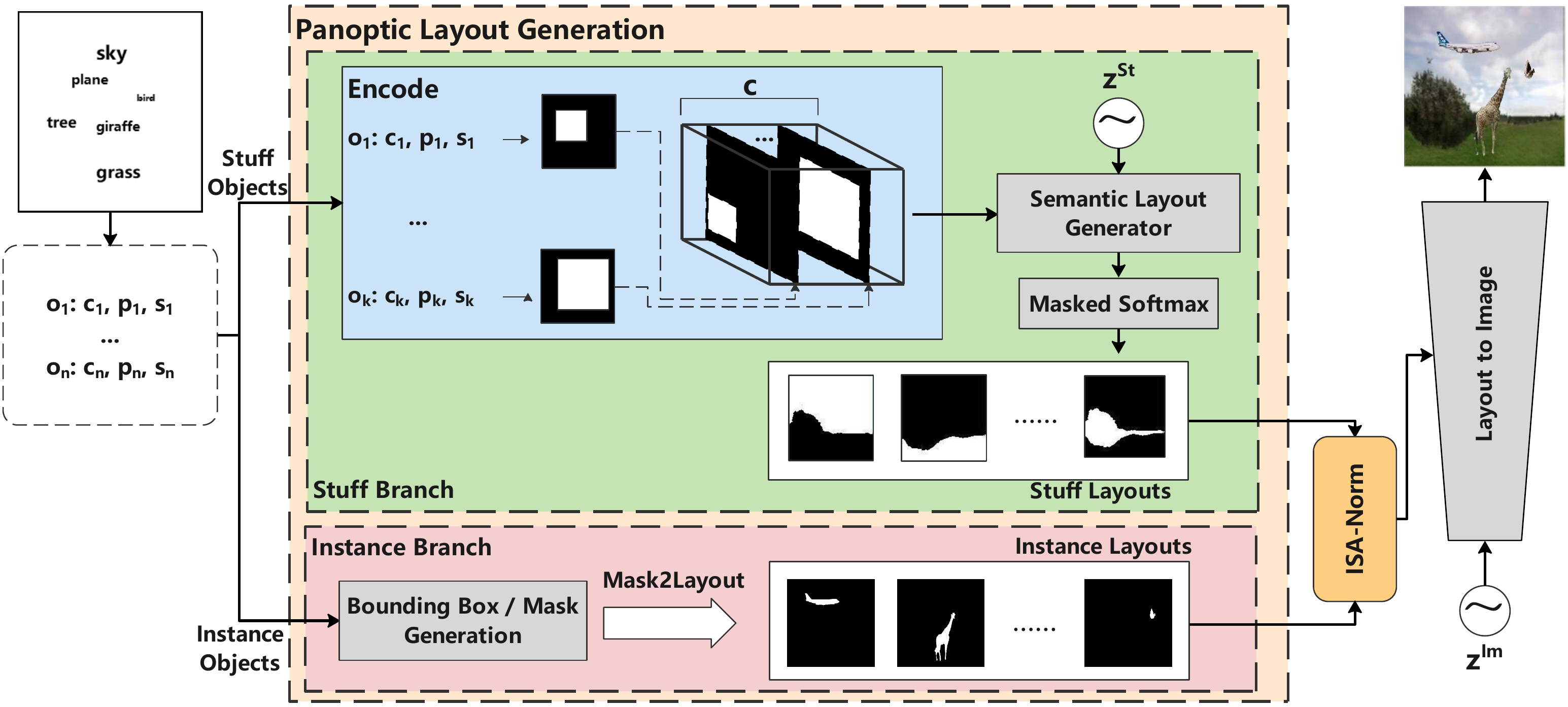}
   \end{center}
   \vspace{-.5cm}
   \caption{Overview of the PLGAN architecture. PLGAN is an interactive image synthesis model trained in an end-to-end manner. It consists of two stages: scene-to-layout generation and layout-to-image synthesis.
      Inspired by panoptic segmentation \cite{kirillov2019panoptic}, panoptic layout generation (PLG) is proposed for scene-to-layout generation, which distinguishes between instances and stuff for generated objects.
      In particular, the stuff layout complements the instance layout, as the latter is prone to the region missing problem.}
   \label{fig:overview}
\end{figure*}

\section{Related Work}
Even since the work of Generative Adversarial Networks (GAN) \cite{NIPS2014_5ca3e9b1} has photo-realistic image synthesis attracted considerable attention.
Further milestones are achieved by BigGAN \cite{brock2018large} and StyleGAN \cite{karras2019style} regarding network architecture and training strategy that bring impressive image quality in high resolution.
While unconditional synthesis models like GAN take random noise as input, conditional synthesis \cite{mirza2014conditional,reed2016generative,zhang2017stackgan,zhu2017unpaired, kim2017learning} takes additional conditions (\eg category, scene, layout) as input to control contents and styles of generated images.
Among these, image synthesis from layout is an effective way for computer-human interaction \cite{sun2019image, li2019object, hong2018inferring}, which requires to assign class label to each pixel.
Layout-to-image models \cite{park2019semantic, wang2018high, zhu2020sean} can produce aesthetically pleasing results with multiple objects by leveraging locations and shapes of objects directly.

\textbf{Instance Layout.} Instance layout based methods consider each object as an instance attached with a bounding box and a shape independently, and assign every pixel with a category label and an instance ID.
For instance, text-to-image model~\cite{hong2018inferring} takes text descriptions as inputs and adopts LSTM~\cite{hochreiter1997long} to predict instance bounding boxes and masks.
The configurable scene layout \cite{johnson2018image, zhao2019image, ashual2019specifying, sun2019image} is a more user-friendly instrument, which consists of a set of objects with labels, locations and their interactions.
SG2Im~\cite{johnson2018image} and Grid2Im~\cite{ashual2019specifying} take advantage of graph convolution networks~\cite{scarselli2008graph} to extract information from scenes and build instance embedding features to predict bounding boxes and masks for layout construction.
LostGAN~\cite{sun2019image} receives bounding boxes and classes as inputs for image synthesis.
Compared to other condition formats, the scene layout provides a similar spatial structure to the target image and is easier to be constructed.
Though the instance-based methods can generate realistic instances with recognizable shapes, they tend to cause region missing problems especially in high-resolution and interactive scenarios where the image lattice cannot be covered entirely by instance masks.

\textbf{Panoptic Layout.}
Our proposed PLGAN leverages panoptic theory from panoptic segmentation.
Panoptic segmention was first termed in \cite{kirillov2019panoptic} to unify instance segmentation and semantic segmentation for scene understanding missions.
Panoptic segmentation models \cite{li2019attention,cheng2020panoptic,li2020unifying} typically utilize separate branches to generate region-based instance layout for \emph{things} and dense-pixel semantic layout for \emph{stuff}.
Inspired by previous works, our PLGAN adopts panoptic layouts in the context of interactive image generation, allowing users to manipulate the entire scene and create photo-realistic images on the fly.

\section{Method} \label{sec:method}
The proposed PLGAN follows a two-stage procedure: \textbf{scene-to-layout generation} and \textbf{layout-to-image synthesis};
see Fig.~\ref{fig:overview} for an overview of its architecture. 
The input of PLGAN is a set of objects $O=\{o_1,o_2,...,o_n\}$ with each $o_i=(c_i, p_i, s_i)$, where $c_i \in \mathcal{C}$ is the object category (\eg $|\mathcal{C}|$ = 171 in the COCO-Stuff dataset), $p_i \in [0,1]^2$ is the center position, and $s_i$ is the object size (typically drawn from some predefined set, \eg $[1, 25]$).

For scene-to-layout construction, we propose the \textbf{Panoptic Layout Generation} (PLG) module inspired from \cite{kirillov2019panoptic}.
Specifically, PLG learns the mapping from scene to panoptic layout by embedding \emph{stuff} and \emph{thing} labels separately:
\begin{align}
L^{St} &= G^{St}(O^{St}, z^{St}),\\
L_i^{Th} &= G^{Th}(o_i^{Th}, z^{Th}_i),
\end{align}
where $L^{St} \in \mathbb{R}^{H\times W \times |\mathcal{C}^{St}|}$ represents the \emph{stuff layout}, $L_i^{Th} \in \mathbb{R}^{H\times W \times 1}$ is the \emph{instance layout}, $G^{St}$ and $G^{Th}$ are two distinctive generators under stuff branch and instance branch respectively, $O^{St}$ and $O^{Th}=\bigcup_i\{o^{Th}_i\}$ are sets of stuff objects and thing objects, and $z^{St},z_i^{Th}\in \mathbb{R}^m$ are latent codes sampled from standard Gaussians.

Meanwhile, layout-to-image synthesis has been well explored in the recent literature. The PLGAN leverages the state-of-the-art models, such as Grid2Im \cite{ashual2019specifying}, LostGAN-V1 \cite{sun2019image}, LostGAN-V2 \cite{sun2020learning} and CAL2I \cite{he2021context}, for layout-to-image synthesis. In mathematical terms, a photorealistic image $I^f$ is produced by the generator $G^{Im}$:
\begin{equation} \label{eq:fakeimage}
I^f = G^{Im}(L^{St}, \{L_i^{Th}\}, z^{Im}),
\end{equation}
from layouts $L^{St},~\{L_i^{Th}\}$ and a Gaussian latent code $z^{Im} \in \mathbb{R}^m$.
Last but not least, we have integrated the \textbf{Instance- and Stuff-Aware Normalization} (ISA-Norm) into the layout-to-image stage, which is dedicated to the fusion of stuff layout and instance layouts.

\subsection{Panoptic Layout Generation}
In PLG we split object categories in $\mathcal{C}$ into two disjoint subsets, \textbf{stuff} $\mathcal{C}^{St}$ and \textbf{things} $\mathcal{C}^{Th}$, \ie $\mathcal{C} = \mathcal{C}^{St} \cup \mathcal{C}^{Th}$. 
The stuff represents amorphous background regions of texture or material, such as grass, sky and road.
In contrast, things are typically countable foreground objects with well-defined shapes, such as people, animals and vehicles.
In order to eliminate mission regions from instance layout-based models \cite{zhao2019image, ashual2019specifying}, we propose to split layout construction into \textbf{instance branch} and \textbf{stuff branch}, in analogy to panoptic segmentation \cite{xiong2019upsnet, li2018learning, kirillov2019panoptic}.
Furthermore, we propose to fuse the stuff- and instance-layouts and then refine the panoptic layout conditioning on instance-to-instance and instance-to-stuff relations.

\subsubsection{Instance Layout Branch}
Similar to previous works \cite{sun2020learning, johnson2018image, ashual2019specifying}, we generate the instance layout in two shots.
First, we predict B(ounding) Box and mask for each instance object with conditional generative models.
Given a thing object $o_i^{Th}=(c_i, p_i, s_i)$ with $c_i\in\mathcal{C}^{Th}$, both the mask- and BBox-generators take word embedding of object label, center position $p_i$ and size $s_i$ as the inputs.
To simplify the model, we predict only the height and width of the BBox and then combine them with the input center position to obtain the final BBox.
In the second step, all masks are resized into specific regions defined by BBoxes.
Based on these BBoxes and masks, we further utilize the Mask2Layout module \cite{sun2020learning} to construct the instance layout $L^{Th} \in \mathbb{R}^{H \times W \times |O^{Th}|}$, where the slice $L^{Th}_{:,:,i}$ corresponds to the predicted mask of object $o_i^{Th}$.

\subsubsection{Stuff Layout Branch}
While instance layouts are generated independently from each other, stuff layouts are intercorrelated and ought to be generated jointly.
More specifically, we first generate a coarse layout $L^{St,init} \in \mathbb{R}^{H \times W \times |\mathcal{C}^{St}|}$, where each slice $L^{St,init}_{:,:,c}$ is a coarse mask of stuff object with label $c\in\mathcal{C}^{St}$.
Given a stuff object $o^{St}=(c, p, s)$, we generate a coarse mask by setting a square mask with $s$ as its height and width and $p$ as its center position (see Fig. \ref{fig:overview}).
Then, we use the conditional generative model with four ResBlocks \cite{park2019semantic} to refine this coarse layout $L^{St,init}$ into a stuff layout $\widehat{L}^{St}$.
Finally, we normalize $\widehat{L}^{St}$ using a \emph{masked softmax}:
\begin{equation} \label{eq:msoftmax}
   {L}^{St}_{h,w,c} = \frac{e^{\widehat{L}^{St}_{h,w,c}}}{\sum_{c' \in \mathcal{C}^{in}}e^{\widehat{L}^{St}_{h,w,c'}}}, ~~c\in\mathcal{C}^{in},
\end{equation}
where $\mathcal{C}^{in} \subset \mathcal{C}^{St}$ contains the input stuff categories.

\subsection{Layout-to-Image Synthesis}

\begin{figure}[t] 
   \begin{center}
  \vspace{-.5cm}
  \includegraphics[width=.9\linewidth]{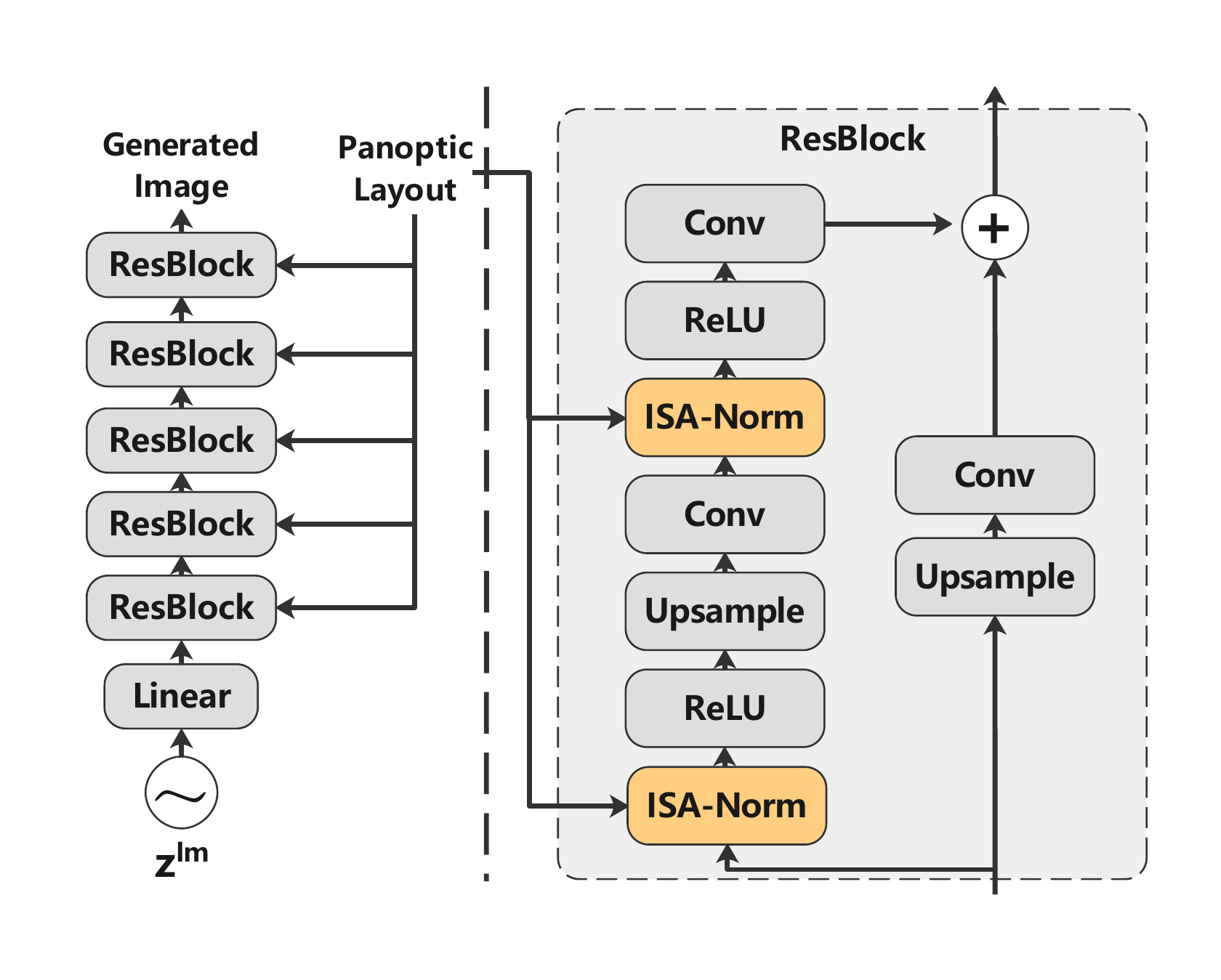}
   \end{center}
   \vspace{-1cm}
    \caption{Illustration of the Layout-to-Image module.}
   \label{fig:generator}
\end{figure}

\subsubsection{Conditional Image Synthesis}
Synthesizing image from layout is a kind of conditional generative task that has recently embarked attention; see, \eg Grid2Im \cite{ashual2019specifying}, LostGAN \cite{sun2020learning} and CAL2I \cite{he2021context}.
To accomplish layout-to-image synthesis in PLGAN, we follow the state-of-the-art method CAL2I \cite{he2021context} to construct our generative model.
LostGAN \cite{sun2020learning} proposes ISLA-Norm which takes instance layout as an input of its conditional generative model.
To fully utilize the panoptic layout in the PLGAN setup, we propose ISA-Norm in place of ISLA-Norm.

Fig.~\ref{fig:generator} shows the design of the layout-to-image stage for $128\times128$ output resolution.
The generator consists of one fully connected (FC) layer and five ResBlocks.
The FC layer maps the image latent code $z^{Im}$ of 128 dimensions to a $4\times4\times128$ tensor.
Then five ResBlocks are employed to successively upsample this tensor to the final generated image of desired resolution, where each ResBlock blends the panoptic layout into upsampling via the ISA-Norm module.

\begin{figure}[t] 
   \begin{center}
  \includegraphics[width=.9\linewidth]{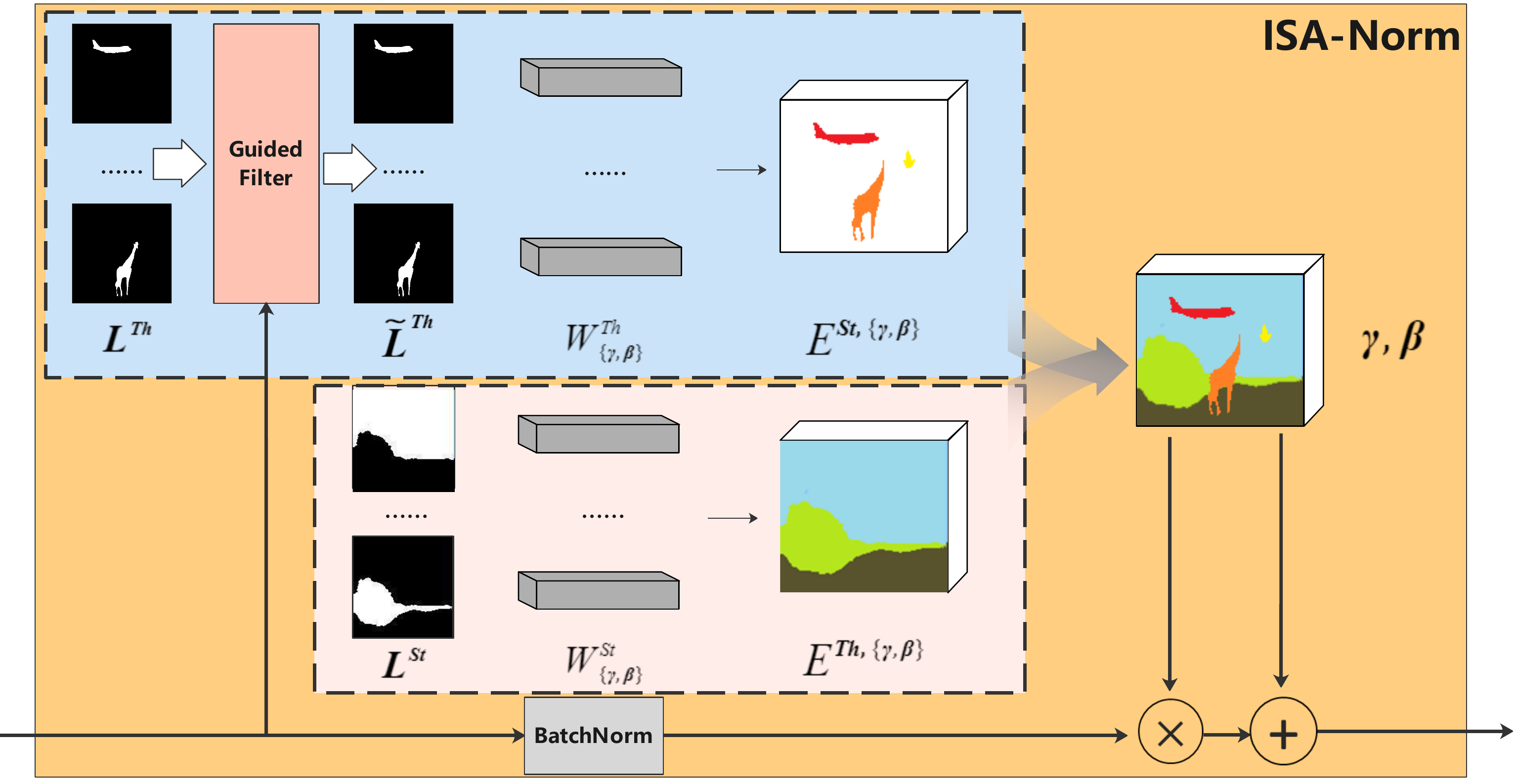}
   \end{center}
   \vspace{-.5cm}
   \caption{Illustration of Instance- and Stuff-Aware Normalization.}
   \label{fig:norm}
\end{figure}

\subsubsection{Instance- and Stuff-Aware Normalization} \label{sec:norm}
Once instance and stuff layout templates are ready, we need to fuse them properly.
In this respect, the ISLA-Norm module was used in LostGAN \cite{sun2020learning} (which only deals with instance layouts).
ISLA-Norm accomplishes multi-object fusion by embedding instance layouts into affine transformations in BatchNorm layers. 
However, direct averaging over embedded instance objects is inappropriate for a panoptic layout.
In a panoptic setup, the stuff layout will overspread the whole scene as background, and among stuff and instances there are widespread overlaps. To address this challenge, we propose \emph{Instance- and Stuff-Aware Normalization} (ISA-Norm); see Fig.~\ref{fig:norm} for an illustration.

Let $X \in \mathbb{R}^{B\times H\times W\times C}$ be the 4D feature map resulting from the activation layer of the ResBlocks in Fig~\ref{fig:generator}.
The ISA-Norm transforms $X$ as in a standard BatchNorm:
\begin{align}
\allowdisplaybreaks
\mu_{c} &= \frac{1}{BHW}\sum_{b,h,w}X_{b,h,w,c}, \\
\sigma_{c} &= \sqrt{\frac{1}{BHW}\Big(\sum_{b,h,w}X_{b,h,w,c}-\mu_{c}\Big)^2 + \epsilon}, \\
\hat{X}_{b,h,w,c} &= \frac{X_{b,h,w,c}-\mu_{c}}{\sigma_{c}} \cdot \gamma_{h,w,c} + \beta_{h,w,c}.
\end{align}
Here $\mu,\sigma\in\mathbb{R}^{C}$ are mean and standard deviation with respect to the batch dimension. The shift- and scale-parameters $\beta,\gamma\in\mathbb{R}^{H\times W\times C}$ are constructed as follows.

First, we use learnable matrices to embed the object classes and get $W^{St}_{\gamma} \in \mathbb{R}^{|\mathcal{C}^{St}|\times C}$ and $W^{Th}_{\beta},W^{Th}_{\gamma} \in \mathbb{R}^{|\mathcal{C}^{Th}|\times C}$ for stuff and thing objects separately.
Then we compute the foreground mask of thing objects:
\begin{equation}
   M_{h,w} = \left\{
   \begin{aligned}
   1, &\quad \textrm{if}~ \sum_c L^{Th}_{h,w,c} > \tau,\\
   0, &\quad \textrm{otherwise},\\
   \end{aligned}
   \right.
\end{equation}
where $\tau$ is a scalar value representing the foreground threshold ($\tau = 0.1$ in experiments).
We further process the instance layouts conditioning on the current image feature via the Guided Filter (GF) \cite{wu2018fast}, and project the instance and stuff layouts into semantic space using label embedding:
\begin{align}
\widetilde{L}_i^{Th} &= \mathrm{GF}(L^{Th}_{i}, X),\\
E^{Th,\star}_{h,w} &= \frac{\left(\sum_i \widetilde{L}_i^{Th}  W^{Th}_{\star} \right)_{h,w}}{\left(\sum_i L^{Th}_{i}\right)_{h,w}} , \\
E^{St,\star} &= L^{St} W^{St}_{\star},
\end{align}
where $\star \in \{\gamma,\beta\}$ and $E^{Th,\star},E^{St,\star} \in \mathbb{R}^{H\times W\times d}$. 
Details on GF are left to the Appendix.
Finally, $\gamma,\beta$ are fused from instance and stuff layout embedding:
\begin{align}
    \gamma_{h,w,c} &= M_{h,w}E^{Th,\gamma}_{h,w,c}  + (1 - M_{h,w}) E^{St,\gamma}_{h,w,c}, \label{eq:gamma}\\
    \beta_{h,w,c} &= M_{h,w}E^{Th,\beta}_{h,w}  + (1 - M_{h,w}) E^{St,\beta}_{h,w,c}. \label{eq:beta}
\end{align}

\begin{figure*}[t]
\begin{center}
{\tabcolsep0.01in
\begin{tabular}{cccccccc}
GT & Pert BBox & \multicolumn{2}{c}{Grid2Im} & \multicolumn{2}{c}{LostGAN-V2} & \multicolumn{2}{c}{Ours} \\

\includegraphics[width=0.12\linewidth]{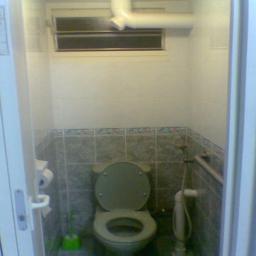}
& \includegraphics[width=0.12\linewidth]{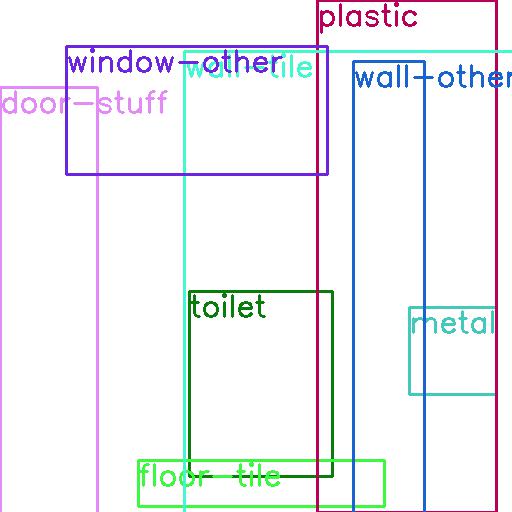}  
& \includegraphics[width=0.12\linewidth]{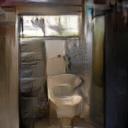}
& \includegraphics[width=0.12\linewidth]{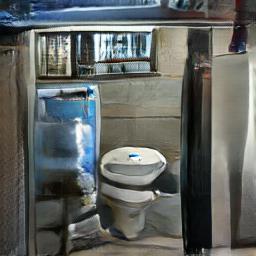}
& \includegraphics[width=0.12\linewidth]{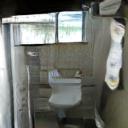}
& \includegraphics[width=0.12\linewidth]{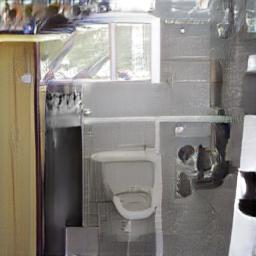}
& \includegraphics[width=0.12\linewidth]{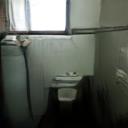}
& \includegraphics[width=0.12\linewidth]{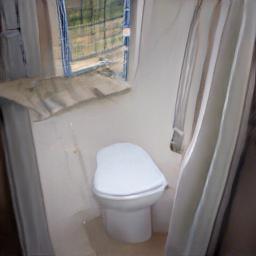} \\

\includegraphics[width=0.12\linewidth]{material/cmp/plgan256/dist_0.3/gt_imgs/000000088462_0.jpg}
& \includegraphics[width=0.12\linewidth]{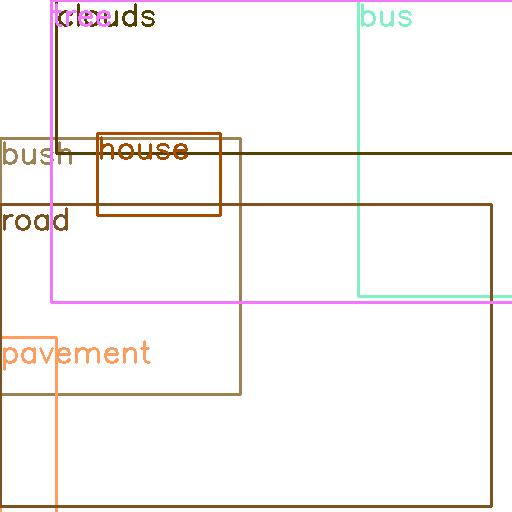}  
& \includegraphics[width=0.12\linewidth]{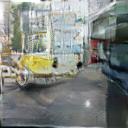}
& \includegraphics[width=0.12\linewidth]{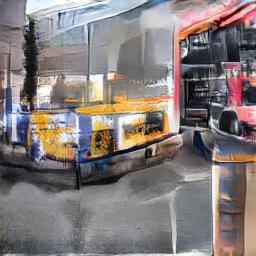}
& \includegraphics[width=0.12\linewidth]{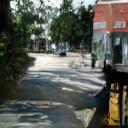}
& \includegraphics[width=0.12\linewidth]{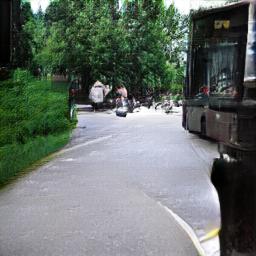}
& \includegraphics[width=0.12\linewidth]{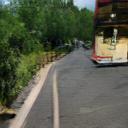}
& \includegraphics[width=0.12\linewidth]{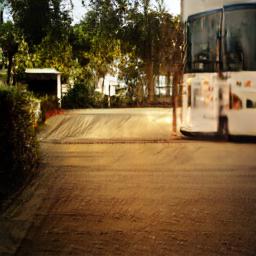} \\

\includegraphics[width=0.12\linewidth]{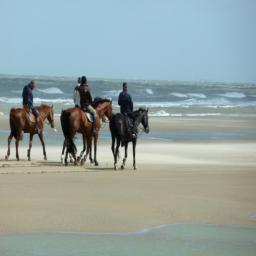}
& \includegraphics[width=0.12\linewidth]{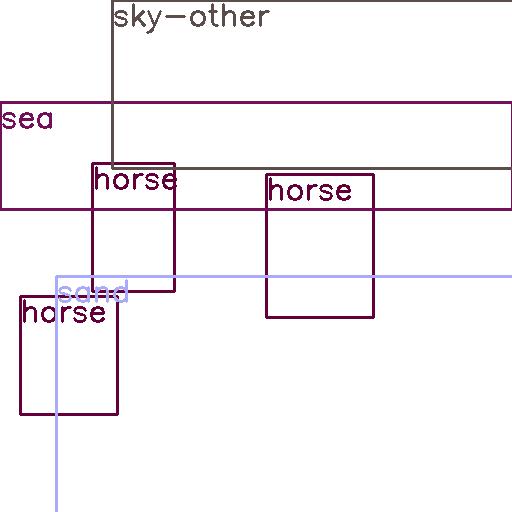}  
& \includegraphics[width=0.12\linewidth]{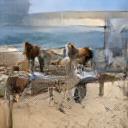}
& \includegraphics[width=0.12\linewidth]{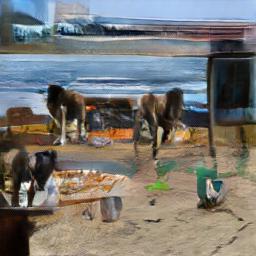}
& \includegraphics[width=0.12\linewidth]{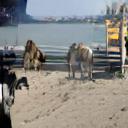}
& \includegraphics[width=0.12\linewidth]{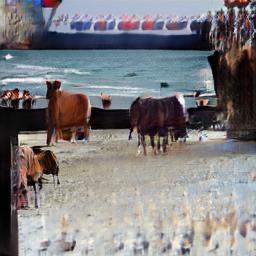}
& \includegraphics[width=0.12\linewidth]{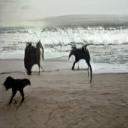}
& \includegraphics[width=0.12\linewidth]{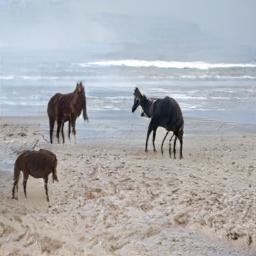} \\
 &  & 128$\times$128 & 256$\times$256 & 128$\times$128 & 256$\times$256 & 128$\times$128 & 256$\times$256
\end{tabular}}
\end{center}
\vspace{-.5cm}
\caption{Visual comparison between sample images generated from perturbed BBoxes (Pert BBoxes) on the COCO-Stuff dataset.}
\label{fig:cmp}
\end{figure*}

\begin{figure*}[t]
 \begin{center}
 {\tabcolsep0.01in
 \begin{tabular}{ccccccccc}

   Input & \multicolumn{4}{c}{Grid2Im} & \multicolumn{4}{c}{Ours} \\

   \includegraphics[width=0.105\linewidth]{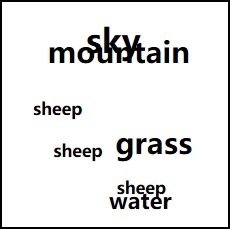}
 & \includegraphics[width=0.105\linewidth]{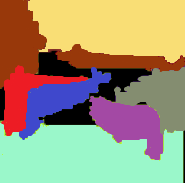}
 & \includegraphics[width=0.105\linewidth]{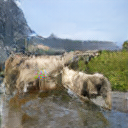}
 & \includegraphics[width=0.105\linewidth]{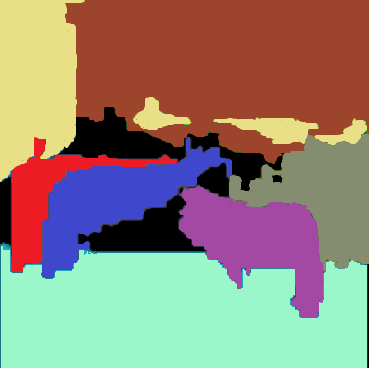}
 & \includegraphics[width=0.105\linewidth]{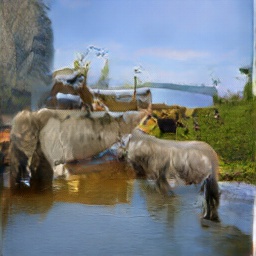}
 & \includegraphics[width=0.105\linewidth]{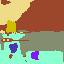}
 & \includegraphics[width=0.105\linewidth]{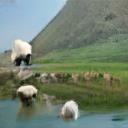}
 & \includegraphics[width=0.105\linewidth]{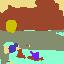}
 & \includegraphics[width=0.105\linewidth]{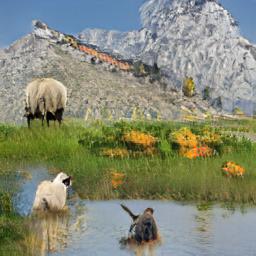}\\


 \includegraphics[width=0.105\linewidth]{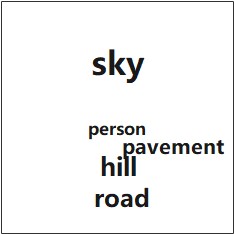}
 & \includegraphics[width=0.105\linewidth]{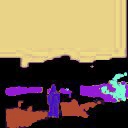}
 & \includegraphics[width=0.105\linewidth]{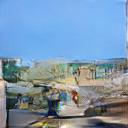}
 & \includegraphics[width=0.105\linewidth]{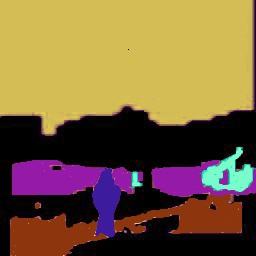}
 & \includegraphics[width=0.105\linewidth]{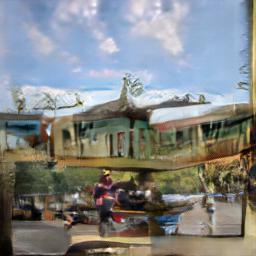}
 & \includegraphics[width=0.105\linewidth]{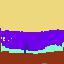}
 & \includegraphics[width=0.105\linewidth]{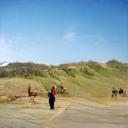}
 & \includegraphics[width=0.105\linewidth]{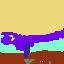}
 & \includegraphics[width=0.105\linewidth]{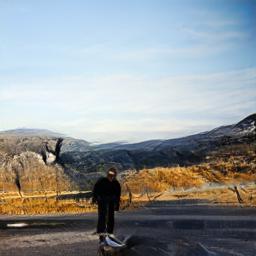}\\

   \includegraphics[width=0.105\linewidth]{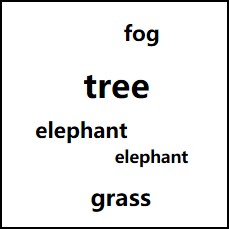}
 & \includegraphics[width=0.105\linewidth]{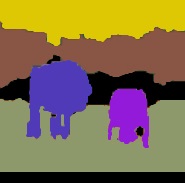}
 & \includegraphics[width=0.105\linewidth]{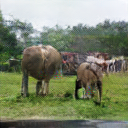}
 & \includegraphics[width=0.105\linewidth]{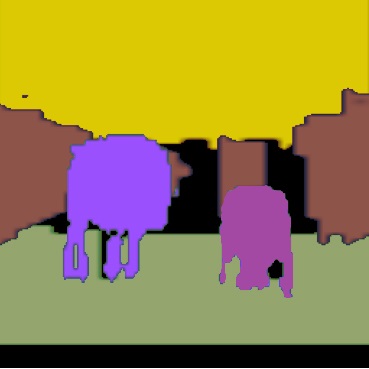}
 & \includegraphics[width=0.105\linewidth]{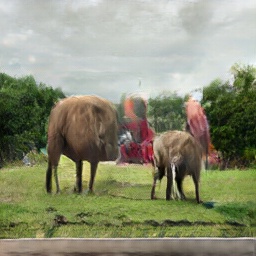}
 & \includegraphics[width=0.105\linewidth]{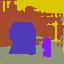}
 & \includegraphics[width=0.105\linewidth]{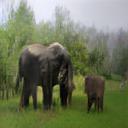}
 & \includegraphics[width=0.105\linewidth]{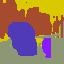}
 & \includegraphics[width=0.105\linewidth]{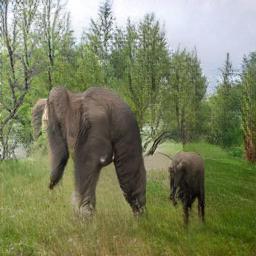}\\

 & \multicolumn{2}{c}{$128\times128$} & \multicolumn{2}{c}{$256\times256$} & \multicolumn{2}{c}{$128\times128$} & \multicolumn{2}{c}{$256\times256$} \\

 \end{tabular}}
 \end{center}
 \vspace{-.5cm}
 \caption{Visual comparison between instance layouts and panoptic layouts on the COCO-Stuff dataset.}
 \label{fig:cmp_layout}
 \end{figure*}

\subsection{Model Objectives} \label{sec:objective}
The total loss used for PLGAN training is given by:
\begin{align} \label{eq:loss}
   \mathcal{L} &= \lambda_{1}\mathcal{L}_{box}+\lambda_{2}\mathcal{L}_{img}+\lambda_{3}\mathcal{L}_{obj} \nonumber\\
               &\quad +\lambda_{4}\mathcal{L}_{per}+\lambda_{5}\mathcal{L}_{rec}+\lambda_{6}\mathcal{L}_{app}. 
\end{align}
Here $\mathcal{L}_{box}$ is the MSE between the predicted bounding boxes and the ground truth bounding boxes,
$\mathcal{L}_{img}$ and $\mathcal{L}_{obj}$ are two \emph{adversarial losses} for images and objects respectively, $\mathcal{L}_{per}$ is the perceptual loss, $\mathcal{L}_{rec}$ is the reconstruction loss, and $\mathcal{L}_{app}$ is another adversarial loss for appearance. 
The balancing weights are set manually in the experiements, \ie $\lambda_{2}= 0.1,~\lambda_{1}=\lambda_{3}=\lambda_{4}=\lambda_{5}=\lambda_{6}=1$. 

\paragraph{Image and object losses.} \hspace{-.2cm}
We use hinge loss \cite{lim2017geometric} for $\mathcal{L}_{img}$:
\begin{align}
\mathcal{L}_{img} &= \mathbb{E}_{I^f\sim p_{fake}}[-D_{img}(I^f)], 
\end{align}
where $I^{r}$ is a real image drawn from training data and $I^{f}$ is a fake image generated from \eqref{eq:fakeimage}. 
Notice that $\mathcal{L}_{img}$ involves a discriminator $D_{img}$ which is updated via minimizing the adversarial loss:
\begin{align}
\mathcal{L}^{adv}_{img}(D_{img}) &= \mathbb{E}_{I^{r}\sim p_{real}}[\max(0,1-D_{img}(I^{r}))] \nonumber \\
    &\quad + \mathbb{E}_{I^{f}\sim p_{fake}}[\max(0,1+D(I^{f}))],
\end{align}
and similarly for the object loss $\mathcal{L}_{obj}$. The adversarial loss has been proven effective in generating realistic textures. 

\paragraph{Reconstruction loss.}
The reconstruction loss measures the $L_1$ distance in pixels between the predicted images and the ground truth images, \ie
\begin{equation}
   \mathcal{L}_{rec}=\mathbb{E}_{I^r\sim p_{real}, I^f\sim p_{fake}} [\|I^{r}-I^f\|_1].
\end{equation}

\paragraph{Perceptual loss.}
The perceptual loss encourages synthesized and real images to share similar feature representations, and is widely used in style transfer \cite{johnson2016perceptual} and image synthesis \cite{zhu2017unpaired}.
With $\phi_j(\cdot)$ the activation of $j$-th layer from the VGG-19 network, the perceptual loss is defined as:
\begin{align}
   \mathcal{L}_{per} &= \mathbb{E}_{\substack{I^r\sim p_{real}\\ I^f\sim p_{fake}}} \Big[ \sum_j w_j\|\phi_j(I^r)-\phi_j(I^f)\|_1 \Big],
\end{align}
with the feature balancing weights $\{w_j\}$. In our experiments, we compute $\mathcal{L}_{per}$ using the activation of \textit{conv1-1}, \textit{conv2-1}, \textit{conv3-1}, \textit{conv4-1} and \textit{conv5} layers, with the corresponding weights $1/32$, $1/16$, $1/8$, $1/4$ and $1$.

\paragraph{Appearance Loss.}
Following CAL2I \cite{he2021context}, we also introduce the appearance loss which penalizes the generator according to pixel-level misalignment:
\begin{align}
\mathcal{L}_{app} &= \mathbb{E}_{I^f\sim p_{fake}}[-D_{app}({A}^f|I^f)], 
\end{align}
The discriminator $D_{app}$ is updated via minimizing the adversarial loss:
\begin{align}
\mathcal{L}_{app}^{adv}(D_{app}) &= \mathbb{E}_{I^r\sim p_{real}}[\max(0, 1 - D_{app}({A}^r|I^r))] \nonumber \\
&\hspace{-.5cm} + \mathbb{E}_{I^f\sim p_{fake}}[\max(0, 1 + D_{app}({A}^f|I^f))].
\end{align}
In $\mathcal{L}_{app}$ and $\mathcal{L}_{app}^{adv}$, we have used ${A}^r$ and ${A}^f$ which are the Gram matrices of object features in ${I}^r$ and ${I}^f$ respectively. The purpose of Gram matrices is to measure the spatial similarity between object features, hence better preserving location-sensitive information in synthesized images \cite{he2021context}.

\begin{table*}[t]
\begin{center}
\caption{Quantitative comparison with respect to Inception Score (higher is better), FID (lower is better) and CAS (higher is better) on the COCO-Stuff dataset. 
Pert1 BBox and Pert2 BBox are generated by perturbing GT BBox with different random biases on object center. 
}
\label{table:cmp_coco}
\setlength{\tabcolsep}{1.0mm}
\begin{tabular}{lcccccccc}
\toprule
Methods     &   Resolution   &\multicolumn{3}{c}{IS$\uparrow$} & \multicolumn{3}{c}{FID$\downarrow$} & {CAS$\uparrow$} \\
\midrule
Real Images & 64$\times$64   & \multicolumn{3}{c}{13.4$\pm$0.5} & \multicolumn{3}{c}{-} &  \multirow{3}{*}{51.04}\\
Real Images & 128$\times$128 & \multicolumn{3}{c}{22.3$\pm$0.4} & \multicolumn{3}{c}{-} &  \\
Real Images & 256$\times$256 & \multicolumn{3}{c}{30.4$\pm$0.6} & \multicolumn{3}{c}{-} & \\
\midrule
  \multicolumn{2}{c}{}                                & GT BBox        & Pert1 BBox   & Pert2 BBox  & GT BBox    & Pert1 BBox & Pert2 BBox & GT BBox \\ \midrule
Layout2Im \cite{zhao2019image} &  \multirow{4}{*}{64$\times$64}  & 9.1$\pm$0.1    & 7.7$\pm$0.2  & 7.0$\pm$0.2 & 37.53      & 44.57      & 50.58  &  27.32 \\ 
Grid2Im \cite{ashual2019specifying} &                 & \textbf{10.3$\pm$0.1}   & -            & -           & 48.7       & -          & -       &  - \\
LostGAN-V1 \cite{sun2019image} &                       & 9.8$\pm$0.2    & -            & -           & 34.31      & -          & -        & 28.81 \\
Ours (CAL2I \cite{he2021context}+PLG) &                  & \textbf{10.3$\pm$0.1}   & \textbf{9.2$\pm$0.1}  & \textbf{8.2$\pm$0.1} & \textbf{21.85}      & \textbf{28.01}      & \textbf{34.52} & \textbf{29.50} \\
\midrule
Grid2Im \cite{ashual2019specifying} & \multirow{8}{*}{128$\times$128} & 11.2$\pm$0.3  & 7.4$\pm$0.1 & 6.4$\pm$0.2  & 63.2       & 77.76     & 87.89       &  25.89   \\
LostGAN-V1 \cite{sun2019image} &                        & 13.8$\pm$0.4  & 9.2$\pm$0.1  & 7.7$\pm$0.1 & 29.65      & 51.96      & 71.04    & 30.68 \\
LostGAN-V2 \cite{sun2020learning} &                     & 14.2$\pm$0.4  & 9.9$\pm$0.1  & 8.1$\pm$0.2 & 24.76      & 43.82      & 59.34    & 31.98 \\
CAL2I \cite{he2021context} &                            & \textbf{15.6$\pm$0.2}  & 11.1$\pm$0.1  & 9.0$\pm$0.1 & 24.15      & 43.12      & 57.89   & 32.52 \\
Grid2Im \cite{ashual2019specifying}+PLG &               & 12.7$\pm$0.1  & 11.0$\pm$0.2 & 9.5$\pm$0.2   & 45.83       & 51.53     & 60.24    &  26.74     \\
LostGAN-V1 \cite{sun2019image}+PLG &                     & 14.1$\pm$0.1  & 12.6$\pm$0.2  & 11.0$\pm$0.2 & 26.85      & 31.82      & 38.67    & 31.33 \\
LostGAN-V2 \cite{sun2020learning}+PLG &                  & 14.6$\pm$0.2  & 12.8$\pm$0.1  & 11.4$\pm$0.1 & 25.43      & 30.80      & 36.75    & 32.86 \\
Ours (CAL2I \cite{he2021context}+PLG) &                   & \textbf{15.6$\pm$0.3}   & \textbf{13.2$\pm$0.2} & \textbf{11.7$\pm$0.2}& \textbf{22.70}      & \textbf{29.03}      & \textbf{35.40}     & \textbf{33.86} \\
\midrule
Grid2Im \cite{ashual2019specifying} & \multirow{3}{*}{256$\times$256}  & 15.2$\pm$0.1  & 7.7$\pm$0.4  & 4.4$\pm$0.1 & 65.95      & 147.85          & 253.59       & 20.54  \\
LostGAN-V2 \cite{sun2020learning} &                     & 18.2$\pm$0.2  & 12.2$\pm$0.2  & 9.5$\pm$0.2  & 30.82      & 56.67      & 77.56        & 30.33 \\
Ours (CAL2I \cite{he2021context}+PLG) &                   & \textbf{18.9$\pm$0.3}  & \textbf{15.8$\pm$0.2}  & \textbf{14.2$\pm$0.2} & \textbf{29.10}      & \textbf{40.14}      & \textbf{46.89}         &  \textbf{32.33} \\
\bottomrule
\end{tabular}
\vspace{-.25cm}
\end{center}
\end{table*}

\subsection{Implementation Details}
The PLGAN model is trained using Pytorch \cite{paszke2019pytorch} on a NVIDIA Tesla V100 GPU server.
The training uses Adam optimizer \cite{kingma2014adam} with learning rate $10^{-4}$ and batch size 128,
and runs for 200 epochs on all tested datasets.
Inference of the PLGAN model is tested on Huawei Atlas inference workstation equipped with Ascend AI Accelerator Card and Ascend Compute Architecture for Neural Networks (CANN).

\section{Experiments}
We evaluate the proposed PLGAN on three datasets: COCO-Stuff \cite{caesar2018coco}, Visual Genome \cite{krishna2017visual}, and our own Landscape dataset.
The results from PLGAN are compared with state-of-the-art methods not only visually and but also quantitatively using widely adopted metrics.
We also carry out ablation studies to evaluate effectiveness of the individual components of PLGAN. Due to space limitation, supplementary results are left to the Appendix.

\textbf{Datasets.}
The \textbf{COCO-Stuff} dataset \cite{caesar2018coco} annotates 40K training image and 5K validation images with bounding boxes and segmentation masks for 80 \textit{thing} categories and 91 \textit{stuff} categories.
Following Ashual \etal \cite{ashual2019specifying}, we choose images with three to eight objects and further filter images whose object coverage is less than 2\%.
We split categories of the \textbf{Visual Genome (VG)} dataset \cite{krishna2017visual} into 92 thing categories and 87 stuff categories. We choose 62,565 training, 5,506 validation and 5,088 test images with 3 to 30 objects in our experiments.
We also created our own \textbf{Landscape} dataset by collecting 27k photos (25k train and 2k val) of $448^2$ resolution from the Flickr website, in order to fully demonstrate the advantage of the stuff layout generation in PLGAN. The Landscape dataset contains only 23 stuff object classes (such as sky, sea and mountain) but no thing objects.
We use a pre-trained UPSNet \cite{xiong2019upsnet} to extract pixel-level segmentation masks for both thing and stuff objects.

\textbf{Methods.}
We compare our PLGAN with state-of-the-art layout-to-image models: 
Layout2Im \cite{zhao2019image}, Grid2Im \cite{ashual2019specifying}, LostGAN-V1 \cite{sun2019image}, LostGAN-V2 \cite{sun2020learning} and CAL2I \cite{he2021context}.
The results from these models are reproduced by the publicly released code.

\textbf{Evaluation Metrics.}
Four metrics are adopted for the quantitative evaluation: Inception Score (\textbf{IS}) \cite{salimans2016improved}, Fr\'echet Inception Distance (\textbf{FID}) \cite{heusel2017gans}, Classification Accuracy Score (\textbf{CAS}), and Coverage (\textbf{COV}) \cite{ivgi2021scene}.
In particular, we compute CAS \cite{ravuri2019classification, sun2020learning} by training a ResNet-101 model on the synthesized images to classify real images for the COCO-Stuff and VG datasets. Higher CAS is better (\ie more identifiable objects). The Coverage (\textbf{COV}) measures the quality of intermediate semantic layouts, which is computed as the average percentage (ranged between $0$ and $100$) of empty area in the generated semantic layout. Higher COV is better (\ie less empty artboard).

\begin{table*}[t]
   \setlength{\tabcolsep}{2.5pt}
   \caption{Ablation study on the PLG and ISA-Norm module.}
   \vspace{-.5cm}
   \begin{center}
   \begin{tabular}{ccccccccccc}
   \toprule
   PLG          &\multicolumn{2}{c}{ISA-Norm}& \multicolumn{3}{c}{IS$\uparrow$}            & \multicolumn{3}{c}{FID$\downarrow$}   & CAS $\uparrow$ & Inference Time\\
                & w/o GF     & w/ GF         & GT BBox       & Pert1 BBox   & Pert2 BBox   & GT BBox    & Pert1 BBox & Pert2 BBox  & GT BBox & \\
   \midrule
                &            &            & 15.6$\pm$0.2  & 11.1$\pm$0.1  & 9.0$\pm$0.1 & 24.15      & 43.12      & 57.89       & 32.52          & 14 ms\\
   \checkmark   &            &            & 14.6$\pm$0.2  & 12.3$\pm$0.1  &10.3$\pm$0.1 & 24.89      & 35.66      & 58.84       & 31.99          & 22 ms\\
   \checkmark   & \checkmark &            & 15.0$\pm$0.3  & 12.5$\pm$0.1  &10.8$\pm$0.1 & 23.65      & 33.09      & 52.73       & 31.09          & 22 ms\\
   \checkmark   &            & \checkmark & 15.6$\pm$0.3  & 13.9$\pm$0.2  &12.8$\pm$0.2 & 22.70      & 27.03      & 33.40       & 33.86          & 26 ms\\
   \bottomrule
   \end{tabular}
   \end{center}
   \label{table:abs}
   \vspace{-.5cm}
\end{table*}

\subsection{Qualitative Results}
In Fig.~\ref{fig:cmp} we show visual comparison between Grid2Im \cite{ashual2019specifying}, LostGAN-V2 \cite{sun2020learning} and our PLGAN under $128^2$ and $256^2$ resolutions, all using perturbed BBoxes as input.
It is clearly observable that Grid2Im and LostGAN-V2 which rely on instance-based layouts produce artifacts in regions between BBoxes and image borders.
This is due to the fact that BBoxes of instances cannot occupy the whole image lattice.
Even if there is no gap between BBoxes, mismatch between instance masks also causes the region missing problem in layouts by Grid2Im.
With increasing resolution, this problem becomes even more apparent.
Our PLGAN adopts panoptic-based layouts and processes stuff objects akin to semantic segmentation,
hence the background of a generated image naturally fills up the whole image lattice.
We refer to Fig.~\ref{fig:cmp_layout} for visual difference between instance layouts and panoptic layouts.

\begin{figure*}[th]
\begin{center}
{\tabcolsep0.005in
\begin{tabular}{ccc}

\includegraphics[height=.23\linewidth]{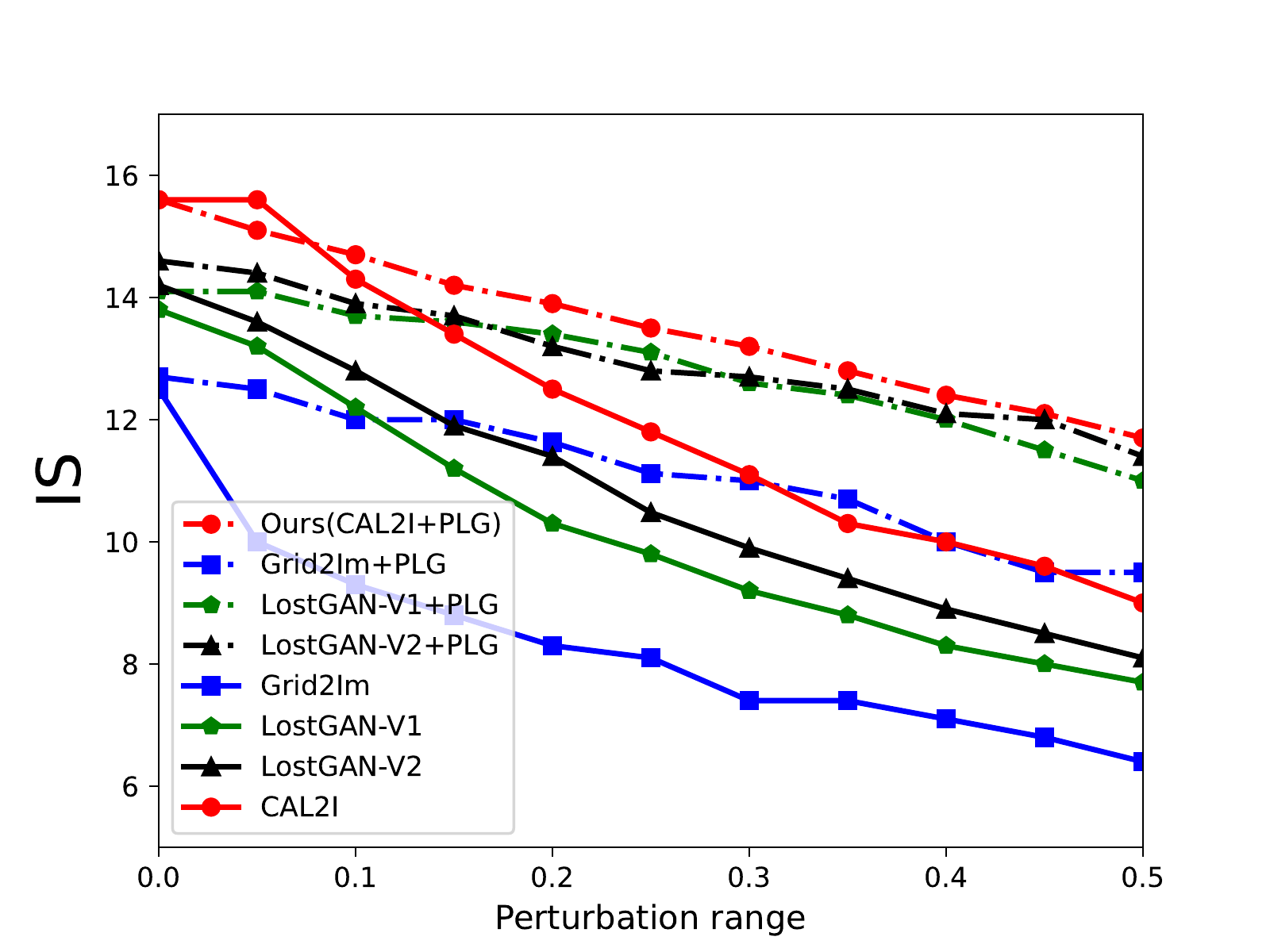} &
\includegraphics[height=.23\linewidth]{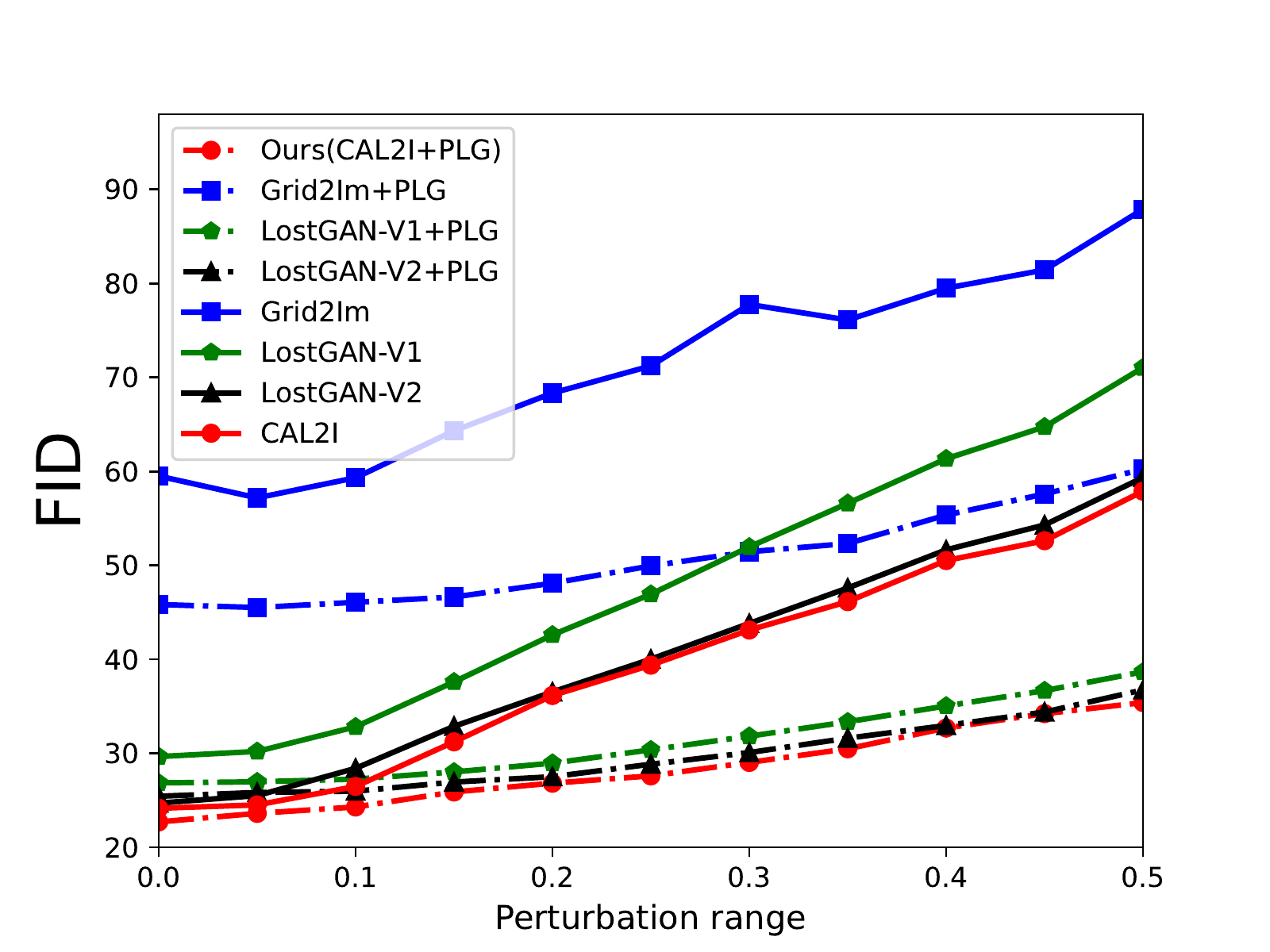} &
\includegraphics[height=.23\linewidth]{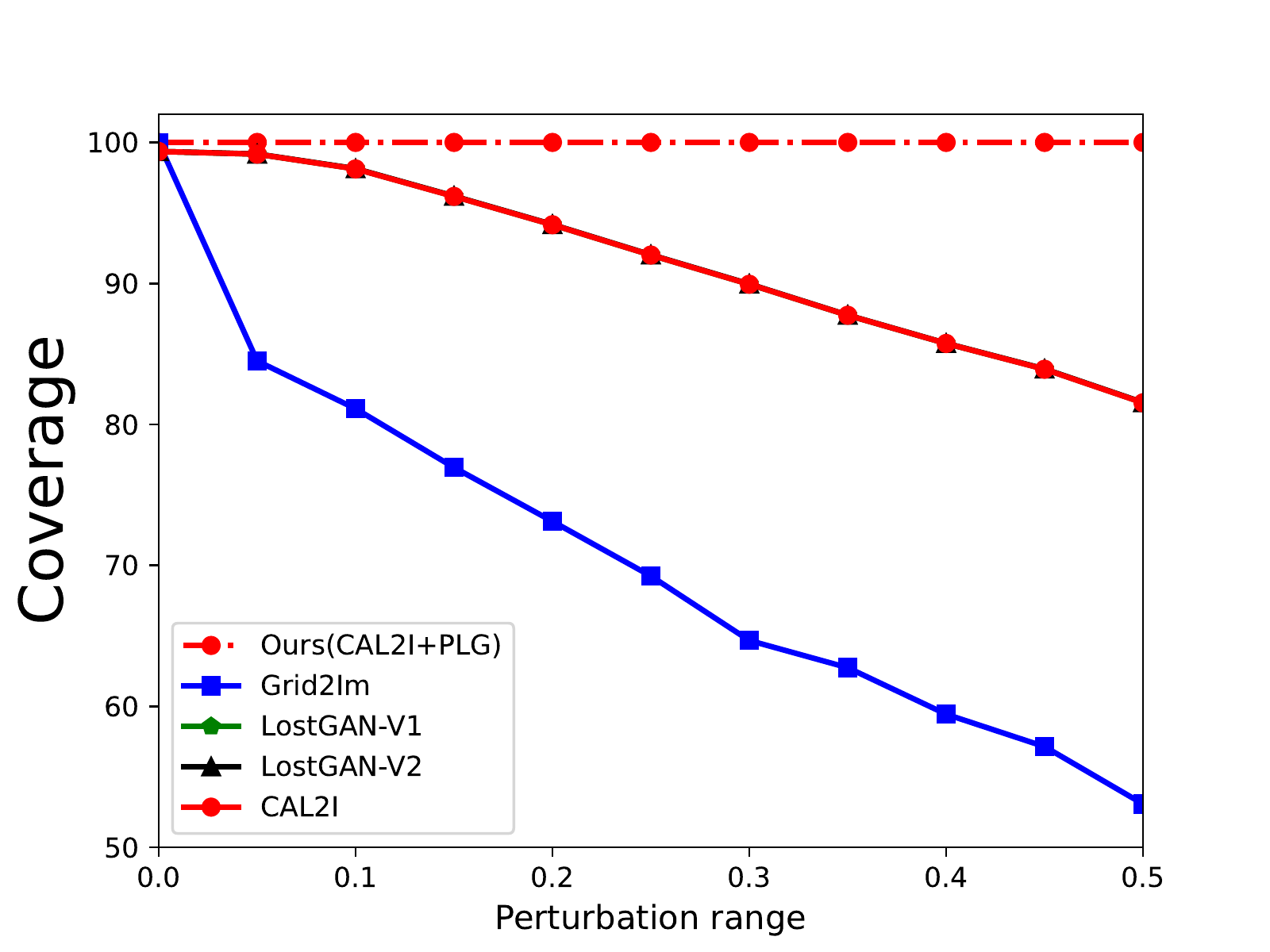}

\end{tabular}}
\end{center}
\vspace{-.5cm}
\caption{IS, FID and Coverage curves with varying perturbation range on the COCO-Stuff dataset under resolution $128\times128$.}
\label{fig:curve_128}
\end{figure*}

\subsection{Quantitative Results}
In Tab.~\ref{table:cmp_coco}, we show quantitative comparisons on COCO-Stuff with respect to Inception Score, FID and CAS across different resolutions.
``GT BBox'' refers to the ground-truth annotations from the original dataset; 
``Pert1 BBox'' and ``Pert2 BBox'' refer to randomly biased center positions of GT BBox in the range $[-0.3, 0.3]$ and $[-0.5, 0.5]$ respectively. 
Unsurprisingly, IS and FID under GT BBox are better than those under Pert1 BBox and Pert2 BBox. 
With GT BBox as input, our PLGAN (CAL2I+PLG) is the best in IS and FID but the difference to the competing models is small. 
However, the advantage of PLGAN in IS and FID is more apparent under Pert1 BBox and Pert2 BBox.
This is due to that perfect placement of BBoxes potentially alleviates the region missing and overlapping problems that previous instance-layout based models are troubles with.
To compare the CAS, we test the synthesized images with GT BBox as input only.
Following LostGAN-V2 \cite{sun2020learning}, we compute CAS on cropped and resized objects at $32^2$ resolution from synthesized and real images.
According to Tab.~\ref{table:cmp_coco}, our methods has higher CAS at different resolutions.
This confirms that, with refined instance layouts by the ISA-Norm module, objects in generated images will attain higher fidelity.

Notably, our PLGAN uses CAL2I as the layout-to-image generator, which can be replaced by other instance-layout based models.
For this reason, we also include results in Tab.~\ref{table:cmp_coco} that combine the PLG module and different layout-to-image generators. Noticeably PLG always improves instance-layout baselines. 

We further elaborate the test on robustness against perturbation on GT BBoxes. By varying the perturbation range from $0$ to $0.5$, Fig.~\ref{fig:curve_128} plots IS, FID and COV curves for our PLGAN and other models from Tab.~\ref{table:cmp_coco}.
It is observed that IS and FID deteriorate on all models as perturbation on GT BBoxes increases.
However, our proposed PLG module always robustifies image synthesis compared to instance-layout baselines. 
The PLGAN which combines CAL2I and PLG is the best overall performer in terms of IS and FID. 
Regarding the coverage metric, we observe visible decay for all instance-layout baselines as the perturbation range increases. Meanwhile, our PLGAN maintains 100\% COV thanks to the panoptic layout.

\begin{table}[t]
   \caption{Ablation study on panoptic layout.}
   \vspace{-.5cm}
   \begin{center}
   \setlength{\tabcolsep}{1.5pt}
   \begin{tabular}{lccc}
   \toprule
   Layout                & IS$\uparrow$  & FID$\downarrow$ & CAS$\uparrow$  \\
   \midrule
   Stuff Layout only     & 12.7$\pm$0.6  & 43.70 & 27.15 \\
   Instance Layout only  & 15.6$\pm$0.2  & 24.15 & 32.52\\
   Panoptic Layout (Instance+Stuff) & 15.6$\pm$0.3  & 22.70 & 33.86 \\
   \bottomrule
   \end{tabular}
   \end{center}
   \vspace{-.5cm}
   \label{table:semantic}
\end{table}

\subsection{Ablation Studies}
We now validate the effectiveness of individual components of the PLGAN. 
In Tab.~\ref{table:abs}, we use CAL2I \cite{he2021context} as baseline and augment it with the PLG and ISA-Norm modules. 
We see that, although PLG resolves the region missing problem from the baseline, it sometimes degrades the image quality measured by the three metrics.
The ISA-Norm is the right remedy to enhance the image quality generated from panoptic layouts.
The PLGAN, with all three components combined, attains the highest score in all metrics and maintain the real time.

We also test the variants of PLGAN using either instance or stuff layout branch alone.
In Tab.~\ref{table:semantic}, we see that treating all objects as stuff or things alone yields inferior metrics to the panoptic layout approach.

\section{Conclusion}
This paper focuses on resolving the region missing problem and improving the robustness of scene-to-image synthesis in interactive scenarios.
To this end, our PLGAN leverages panoptic theory and constructs instance and stuff layouts through separate branches.
The resulting panoptic layouts eliminate missing regions and yield aesthetically pleasing images, even if perturbation on object locations is allowed.
Extensive evidences on the COCO-Stuff, Visual Genome, and Landscape datasets advocate the superiority of PLGAN over the state-of-the-art methods.

\section{Acknowledgement}
We gratefully acknowledge the optimization support of Ascend CANN (Compute Architecture for Neural Networks) used for this research.

{\small
\bibliographystyle{ieee_fullname}
\bibliography{egbib}
}

\clearpage
\renewcommand\thesection{\Alph{section}}
\setcounter{section}{0} 

\section{Supplementary Results}

\subsection{Qualitative Results on Visual Genome}
\begin{figure*}[t]
 \begin{center}
 {\tabcolsep0.01in
 \begin{tabular}{ccccc}
 GT & Pert BBox & LostGAN-V1 & CAL2I & Ours (CAL2I+PLG) \\

 \includegraphics[width=0.19\linewidth]{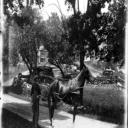}
 & \includegraphics[width=0.19\linewidth]{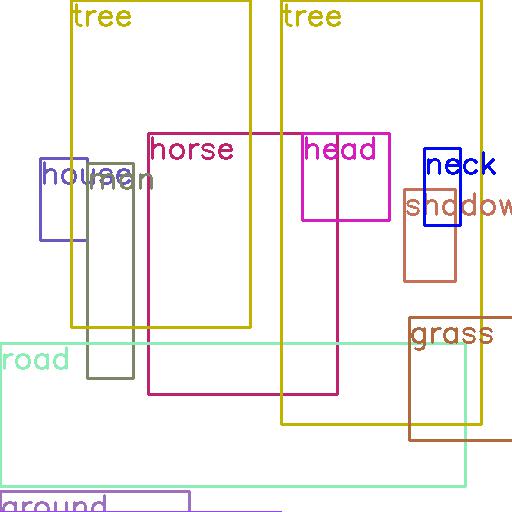}  
 & \includegraphics[width=0.19\linewidth]{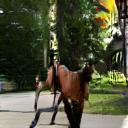}
 & \includegraphics[width=0.19\linewidth]{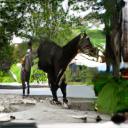}
 & \includegraphics[width=0.19\linewidth]{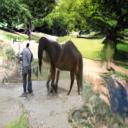} \\

 \includegraphics[width=0.19\linewidth]{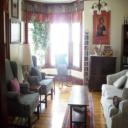}
 & \includegraphics[width=0.19\linewidth]{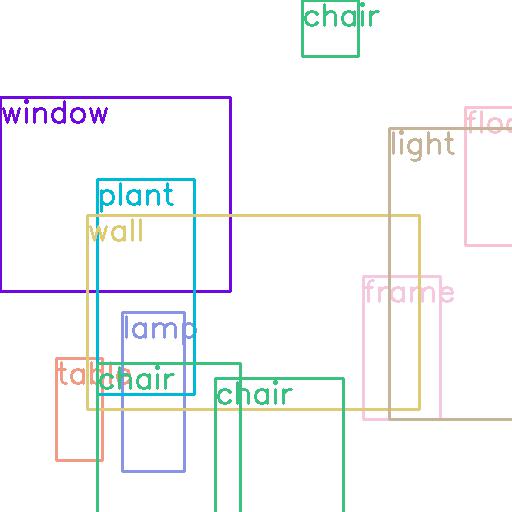}  
 & \includegraphics[width=0.19\linewidth]{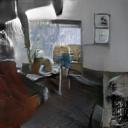}
 & \includegraphics[width=0.19\linewidth]{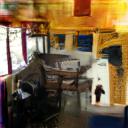}
 & \includegraphics[width=0.19\linewidth]{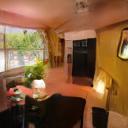} \\

 \includegraphics[width=0.19\linewidth]{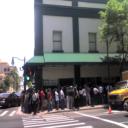}
 & \includegraphics[width=0.19\linewidth]{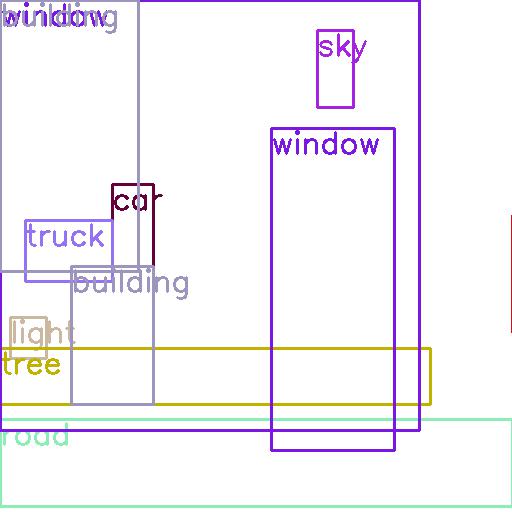}  
 & \includegraphics[width=0.19\linewidth]{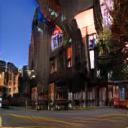}
 & \includegraphics[width=0.19\linewidth]{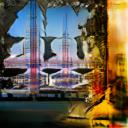}
 & \includegraphics[width=0.19\linewidth]{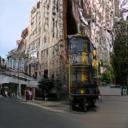} \\

 \includegraphics[width=0.19\linewidth]{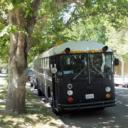}
 & \includegraphics[width=0.19\linewidth]{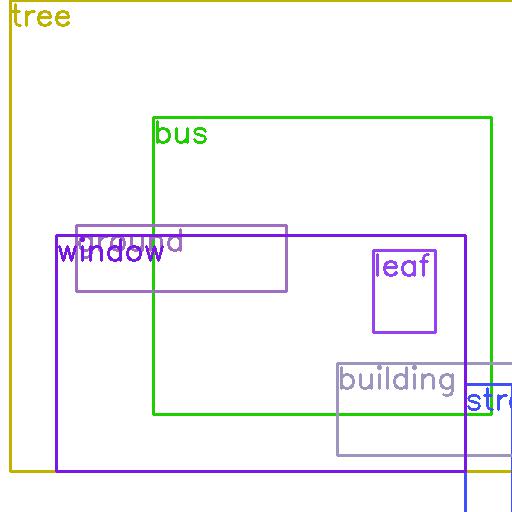}  
 & \includegraphics[width=0.19\linewidth]{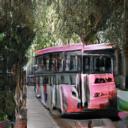}
 & \includegraphics[width=0.19\linewidth]{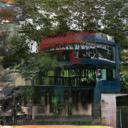}
 & \includegraphics[width=0.19\linewidth]{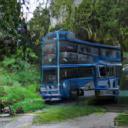} \\

 \includegraphics[width=0.19\linewidth]{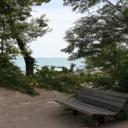}
 & \includegraphics[width=0.19\linewidth]{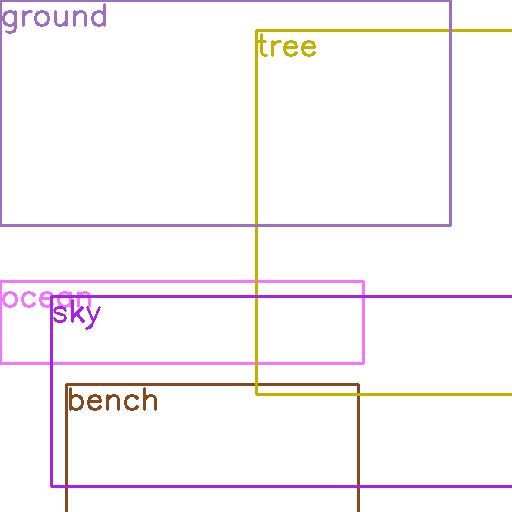}  
 & \includegraphics[width=0.19\linewidth]{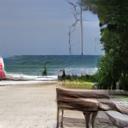}
 & \includegraphics[width=0.19\linewidth]{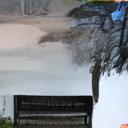}
 & \includegraphics[width=0.19\linewidth]{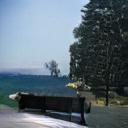} \\

 \includegraphics[width=0.19\linewidth]{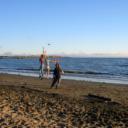}
 & \includegraphics[width=0.19\linewidth]{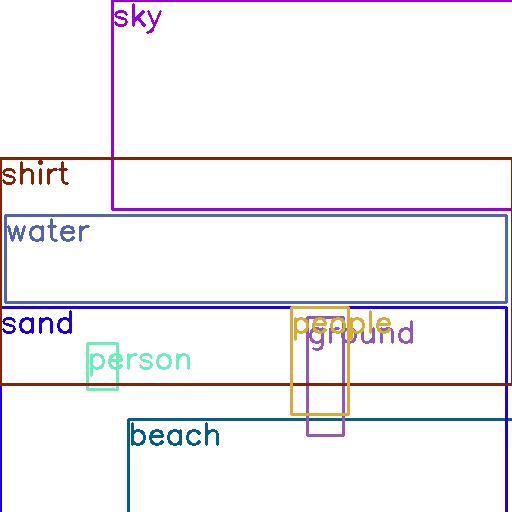}  
 & \includegraphics[width=0.19\linewidth]{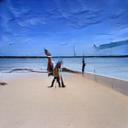}
 & \includegraphics[width=0.19\linewidth]{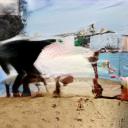}
 & \includegraphics[width=0.19\linewidth]{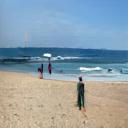} \\

 \end{tabular}}
 \end{center}
 \vspace{-.5cm}
 \caption{Visual comparison between sample images generated from perturbed BBoxes (Pert BBoxes) on the VG dataset.}
 \label{fig:cmp_vg}
 \end{figure*}

\begin{figure*}[t]
 \begin{center}
 {\tabcolsep0.01in
 \begin{tabular}{ccccc}

   \includegraphics[width=0.19\linewidth]{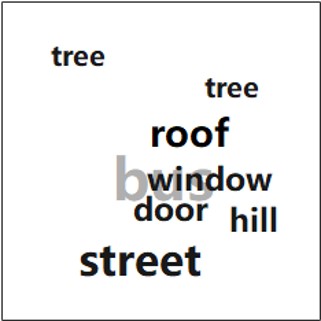}
 & \includegraphics[width=0.19\linewidth]{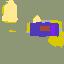}
 & \includegraphics[width=0.19\linewidth]{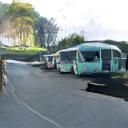}
 & \includegraphics[width=0.19\linewidth]{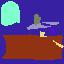}
 & \includegraphics[width=0.19\linewidth]{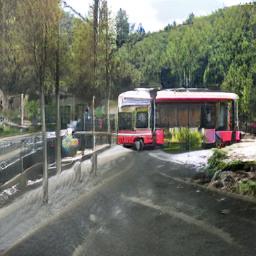}\\

 \includegraphics[width=0.19\linewidth]{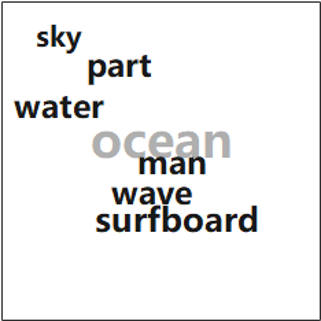}
 & \includegraphics[width=0.19\linewidth]{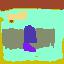}
 & \includegraphics[width=0.19\linewidth]{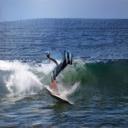}
 & \includegraphics[width=0.19\linewidth]{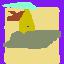}
 & \includegraphics[width=0.19\linewidth]{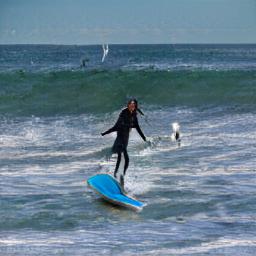}\\

 \includegraphics[width=0.19\linewidth]{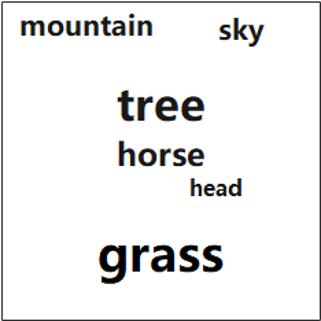}
 & \includegraphics[width=0.19\linewidth]{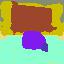}
 & \includegraphics[width=0.19\linewidth]{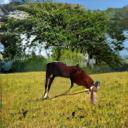}
 & \includegraphics[width=0.19\linewidth]{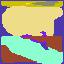}
 & \includegraphics[width=0.19\linewidth]{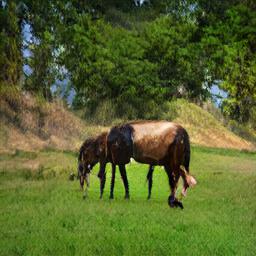}\\

 \includegraphics[width=0.19\linewidth]{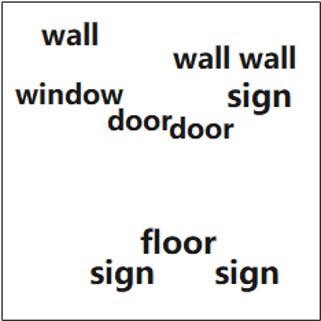}
 & \includegraphics[width=0.19\linewidth]{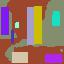}
 & \includegraphics[width=0.19\linewidth]{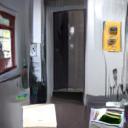}
 & \includegraphics[width=0.19\linewidth]{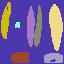}
 & \includegraphics[width=0.19\linewidth]{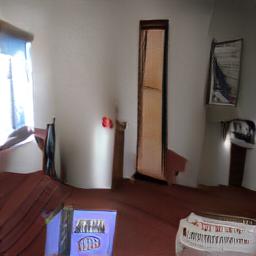}\\

 \includegraphics[width=0.19\linewidth]{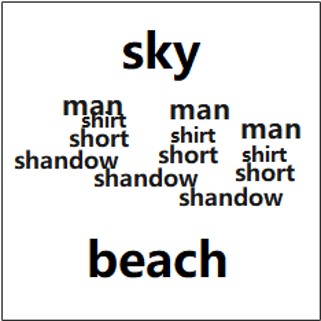}
 & \includegraphics[width=0.19\linewidth]{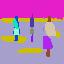}
 & \includegraphics[width=0.19\linewidth]{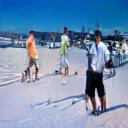}
 & \includegraphics[width=0.19\linewidth]{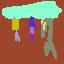}
 & \includegraphics[width=0.19\linewidth]{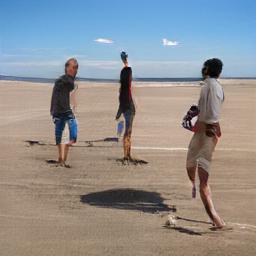}\\

 \includegraphics[width=0.19\linewidth]{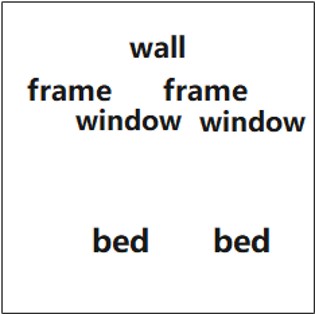}
 & \includegraphics[width=0.19\linewidth]{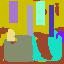}
 & \includegraphics[width=0.19\linewidth]{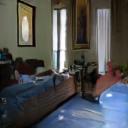}
 & \includegraphics[width=0.19\linewidth]{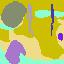}
 & \includegraphics[width=0.19\linewidth]{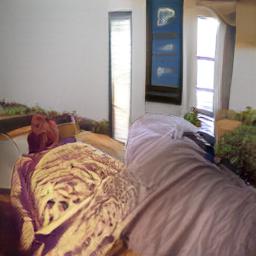}\\

 & \multicolumn{2}{c}{$128\times128$} & \multicolumn{2}{c}{$256\times256$} \\
 Input & Panoptic layout & Generated Image & Panoptic layout & Generated Image \\

 \end{tabular}}
 \end{center}
 \vspace{-.5cm}
 \caption{Synthesized image samples on the VG dataset.}
 \label{fig:cmp_layout_vg}
 \end{figure*}

In Fig.~\ref{fig:cmp_vg}, we show visual comparisons of LostGAN-V1 \cite{sun2019image}, CAL2I \cite{he2021context}, and the proposed PLGAN using perturbed Bounding Boxes as input based on the VG dataset~\cite{krishna2017visual}.
In Fig.~\ref{fig:cmp_layout_vg}, we show sythesized image samples with the corresponding panoptic layouts on the VG dataset under $128^2$ and $256^2$ resolutions.

\subsection{Quantitative Results on Visual Genome}
Similar to Tab.~\ref{table:cmp_coco} in the main paper, Tab.~\ref{table:cmp_vg} reports quantitative comparison with respect to Inception Score, FID and CAS on the VG dataset.

\begin{table*}[ht]
\begin{center}
\caption{Quantitative comparison with respect to Inception Score, FID and CAS on the VG dataset.
}
\label{table:cmp_vg}
\setlength{\tabcolsep}{1.0mm}
\begin{tabular}{lcccccccc}
\toprule
Methods     &   Resolution   &\multicolumn{3}{c}{IS$\uparrow$} & \multicolumn{3}{c}{FID$\downarrow$} & {CAS$\uparrow$} \\
\midrule
Real Images & 128$\times$128 & \multicolumn{3}{c}{20.5$\pm$1.5} & \multicolumn{3}{c}{-} &  \multirow{2}{*}{48.07}\\
Real Images & 256$\times$256 & \multicolumn{3}{c}{28.6$\pm$1.1} & \multicolumn{3}{c}{-} & \\
\midrule
  \multicolumn{2}{c}{}                                & GT BBox        & Pert1 BBox   & Pert2 BBox  & GT BBox    & Pert1 BBox & Pert2 BBox & GT BBox \\ \midrule
LostGAN-V1~\cite{sun2019image} & \multirow{4}{*}{128$\times$128}& 11.1$\pm$0.6  & 10.3$\pm$0.1  & 9.7$\pm$0.1 & 29.36      & 39.48      & 42.29    & 28.85 \\
LostGAN-V2~\cite{sun2020learning} &                     & 10.7$\pm$0.2  & -  & - & 29.00      & -      & -    & 29.35 \\
CAL2I~\cite{he2021context} &                            & 12.6$\pm$0.4  & 8.4$\pm$0.1  & 7.3$\pm$0.1 & 21.78      & 49.53      & 61.30   & 29.2 \\
Ours(CAL2I~\cite{he2021context}+PLG) &                   & \textbf{12.7$\pm$0.2}   & \textbf{10.6$\pm$0.1} & \textbf{10.1$\pm$0.1} & \textbf{20.62}      & \textbf{32.93}      & \textbf{37.03}     & \textbf{30.81} \\
\midrule
LostGAN-V2~\cite{sun2020learning} & \multirow{2}{*}{256$\times$256}  & 14.1$\pm$0.3  & -  & -  & 47.62      & -      & -        & 28.81 \\
Ours(CAL2I~\cite{he2021context}+PLG) &                   & \textbf{14.9$\pm$0.1}  & 13.2$\pm$0.2  & 12.6$\pm$0.1 & \textbf{28.06}      & 38.41     & 41.36      & \textbf{29.35} \\
\bottomrule
\end{tabular}
\end{center}
\end{table*}

\subsection{Qualitative Results on Landscape}
\begin{figure*}[ht]
 \begin{center}
 {\tabcolsep0.01in
 \begin{tabular}{cccccc}

 \rotatebox{90}{\quad \quad \quad Input}
 \includegraphics[width=0.19\linewidth]{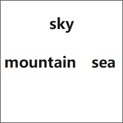}
 & \includegraphics[width=0.19\linewidth]{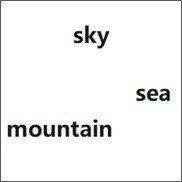}
 & \includegraphics[width=0.19\linewidth]{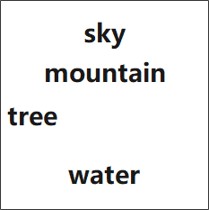}
 & \includegraphics[width=0.19\linewidth]{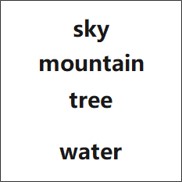}
 & \includegraphics[width=0.19\linewidth]{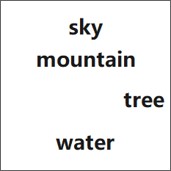}\\

 \rotatebox{90}{\quad \quad \quad Grid2Im}
 \includegraphics[width=0.19\linewidth]{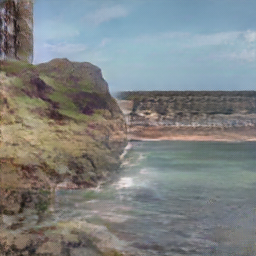}
 & \includegraphics[width=0.19\linewidth]{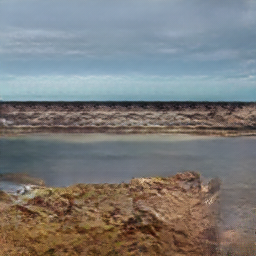}
 & \includegraphics[width=0.19\linewidth]{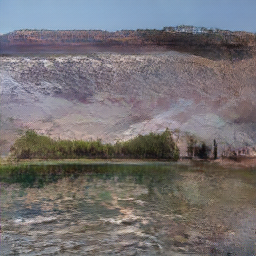}
 & \includegraphics[width=0.19\linewidth]{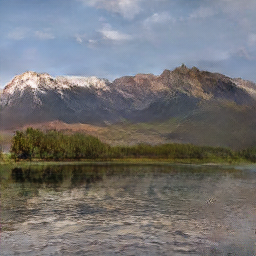}
 & \includegraphics[width=0.19\linewidth]{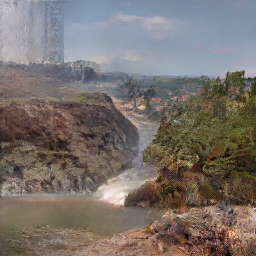}\\

 \rotatebox{90}{\quad \quad \quad Ours}
 \includegraphics[width=0.19\linewidth]{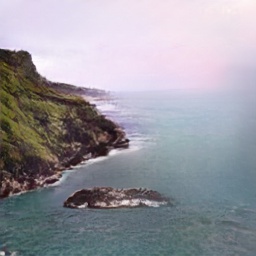}
 & \includegraphics[width=0.19\linewidth]{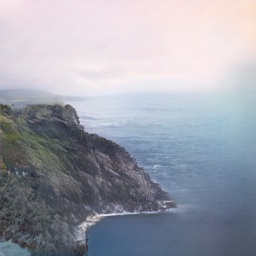}
 & \includegraphics[width=0.19\linewidth]{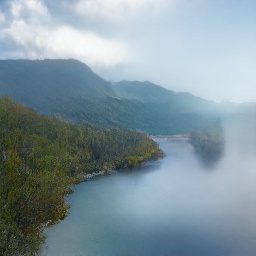}
 & \includegraphics[width=0.19\linewidth]{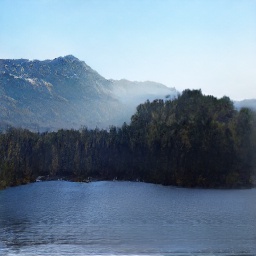}
 & \includegraphics[width=0.19\linewidth]{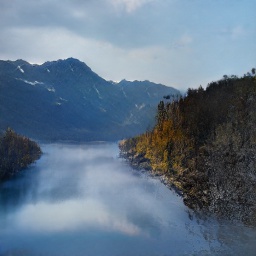}\\

 \end{tabular}}
 \end{center}
 \vspace{-.5cm}
 \caption{Visual comparison on the Landscape dataset.}
 \label{fig:cmp_landscape}
 \end{figure*}

In Fig.~\ref{fig:cmp_landscape}, we compare generated images from Grid2Im \cite{ashual2019specifying} and our PLGAN on the Landscape dataset. 

\subsection{Quantitative Results on Landscape}
In Tab.~\ref{table:abs_landscape}, we quantitavely compare Grid2Im \cite{ashual2019specifying} and our PLGAN on the Landscape dataset,
for which all objects are ``stuff''.
Our method outperforms Grid2Im on all metrics.
The fact that this dataset contains only stuff objects makes the difference even more apparent.

\begin{table}[h]
   \setlength{\tabcolsep}{2.5pt}
   \caption{Quantitative comparison on the Landscape dataset.}
   \begin{center}
   \vspace{-.5cm}
   \begin{tabular}{lccc}
   \toprule
   Method                                & Resolution                    & IS$\uparrow$ & FID $\downarrow$ \\
   \midrule
   Real Images                           & \multirow{3}{*}{448$\times$448} & 5.9$\pm$0.2   & - \\
   Grid2Im \cite{ashual2019specifying}   &                               & 1.8$\pm$0.1   & 144.84\\
   Ours                                  &                               & 3.3$\pm$0.1   & 57.40\\
   \bottomrule
   \end{tabular}
   \end{center}
   \label{table:abs_landscape}
\end{table}

\subsection{Robustness to Perturbed BBoxes}

\begin{figure*}[ht]
\begin{center}
{\tabcolsep0.005in
\begin{tabular}{ccc}

\includegraphics[height=.235\linewidth]{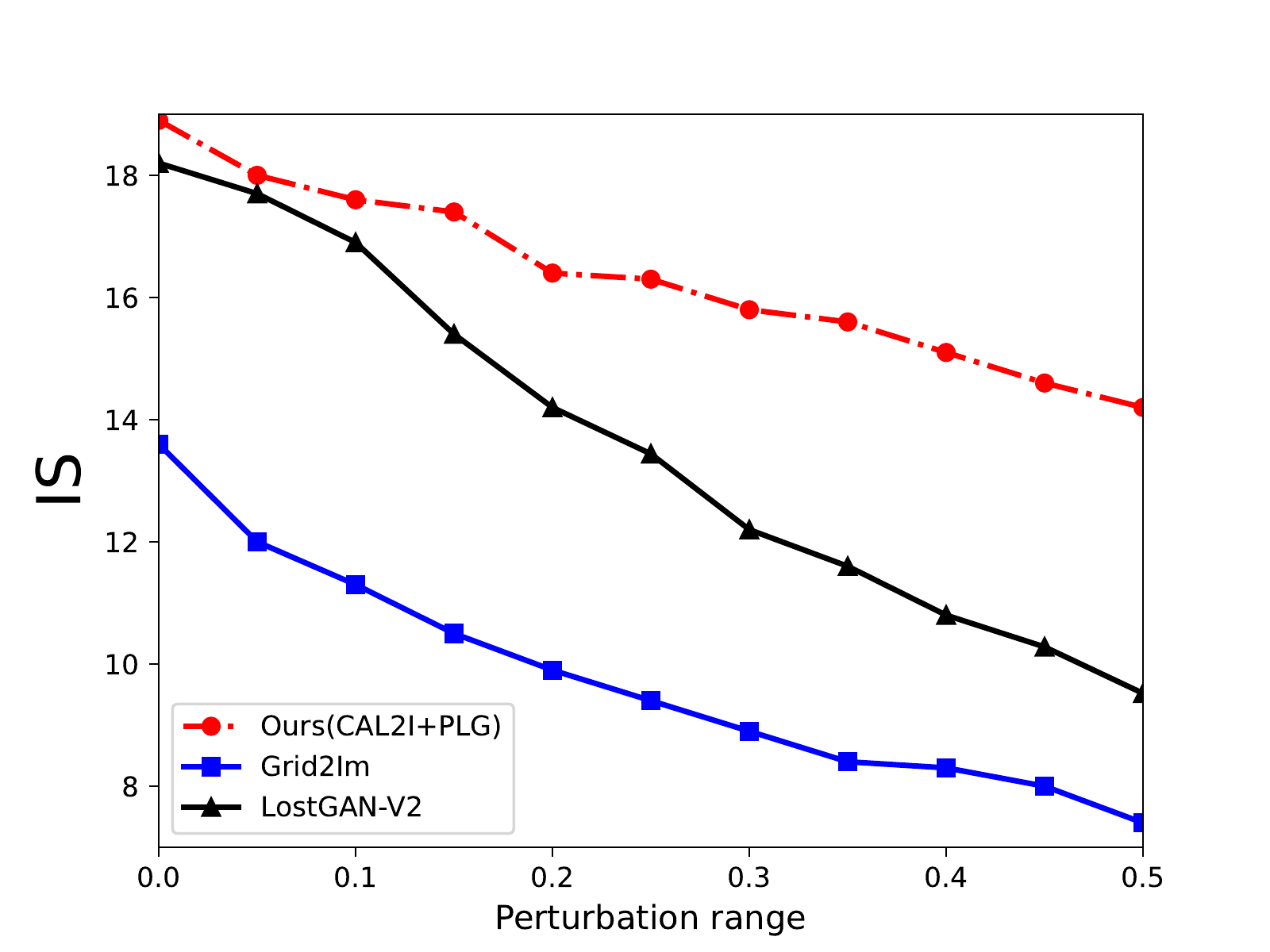}&
\includegraphics[height=.235\linewidth]{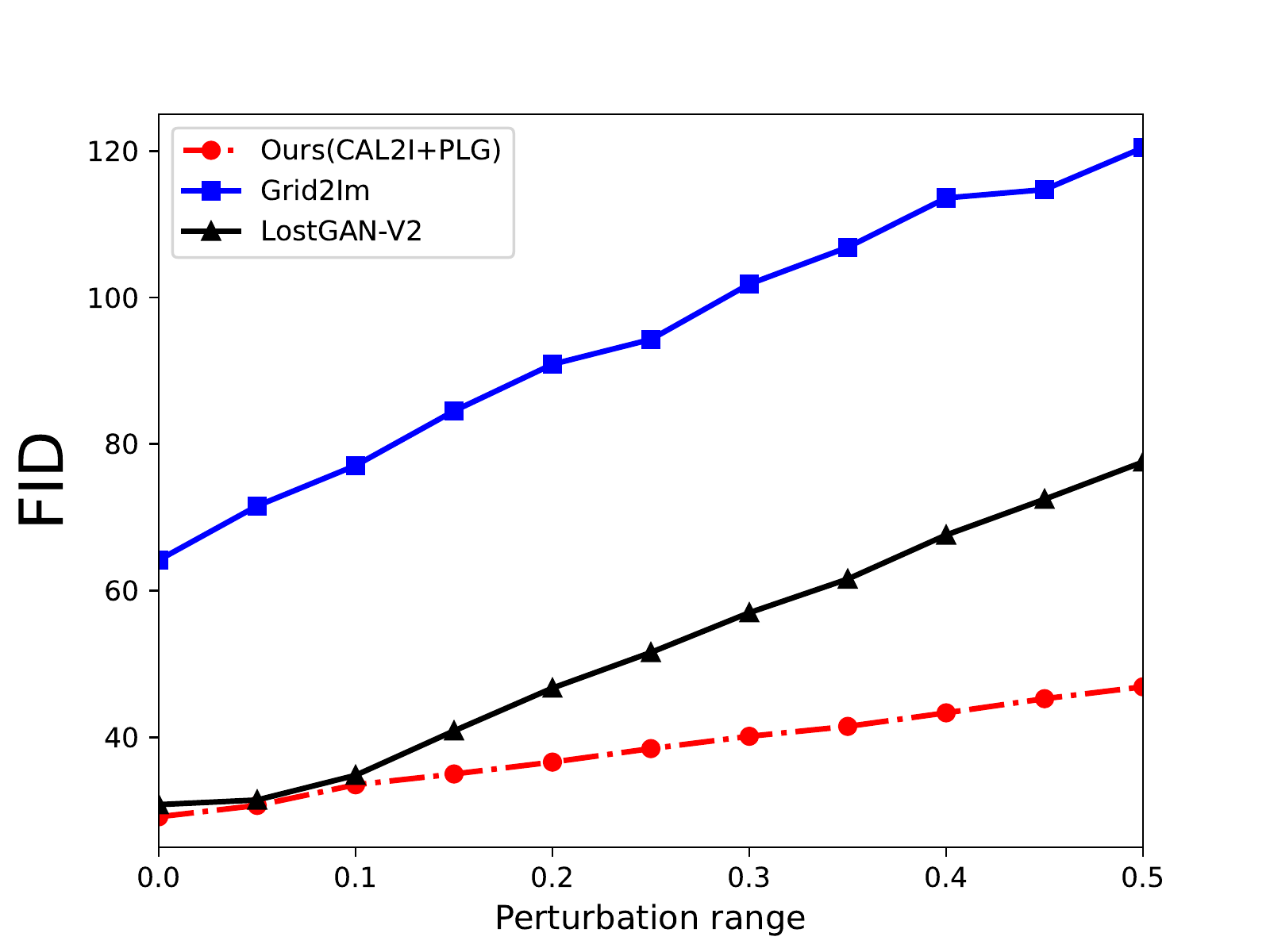} &
\includegraphics[height=.235\linewidth]{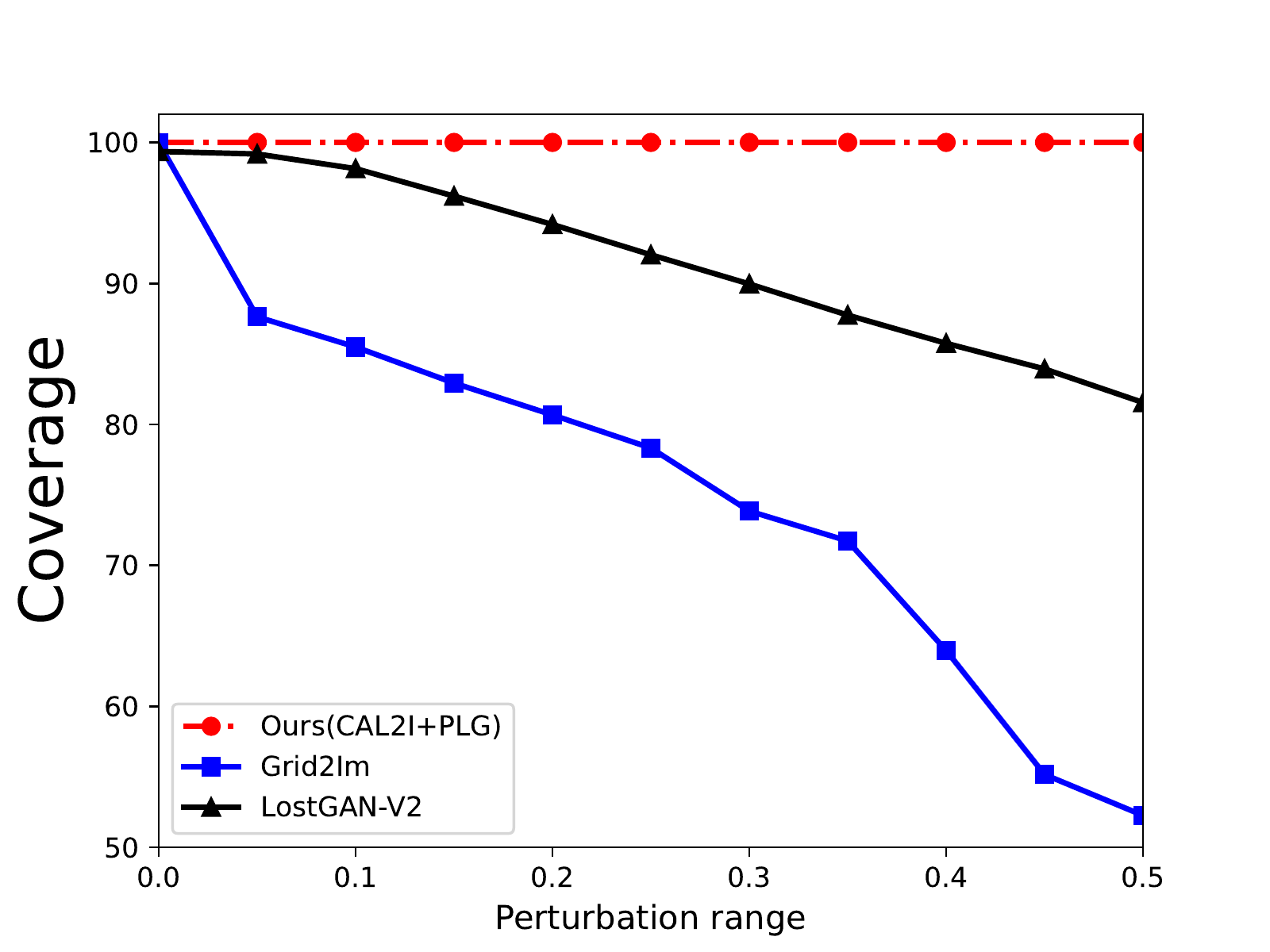}

\end{tabular}}
\end{center}
\vspace{-.5cm}
\caption{IS, FID and Coverage curves with varying perturbation range on the COCO-Stuff dataset under $256\times256$ resolution.}
\label{fig:curve_256}
\end{figure*}

In Fig.~\ref{fig:curve_256}, we plot IS, FID and Coverage curves with varying perturbation range for Grid2Im \cite{ashual2019specifying}, LostGAN-V2 \cite{sun2020learning} and our PLGAN under $256^2$ resolution. Similar to the robustness test under $128^2$ resolution in Fig.~\ref{fig:curve_128}, PLGAN again claims the most robust model among others.

\subsection{User Study}
We conduct a user study on Wjx (https://www.wjx.cn) to rate realism of generated images.
Specifically, we select 100 grouped image samples generated from Grid2Im, LostGAN, CAL2I and our method under $128\times128$ resolution.
Each vote picks one of the two images from the same group and counts one point for the corresponding image generator.
The overall scores after 600 votes in total are shown in Tab.~\ref{table:user_study}.

\begin{table}[h]
   \setlength{\tabcolsep}{2.5pt}
   \caption{User Study statistical results.}
   \begin{center}
   \vspace{-.5cm}
   \begin{tabular}{cccc}
   \toprule
   Grid2Im & LostGAN & CAL2I & PLGAN(ours) \\
   \midrule
   118 & 142 & 143 & 197 \\
   \bottomrule
   \end{tabular}
   \end{center}
   \label{table:user_study}
\end{table}




\section{Guide Filter}
\label{sec:gf}
\begin{figure}[h] 
   \begin{center}
  \includegraphics[width=.9\linewidth]{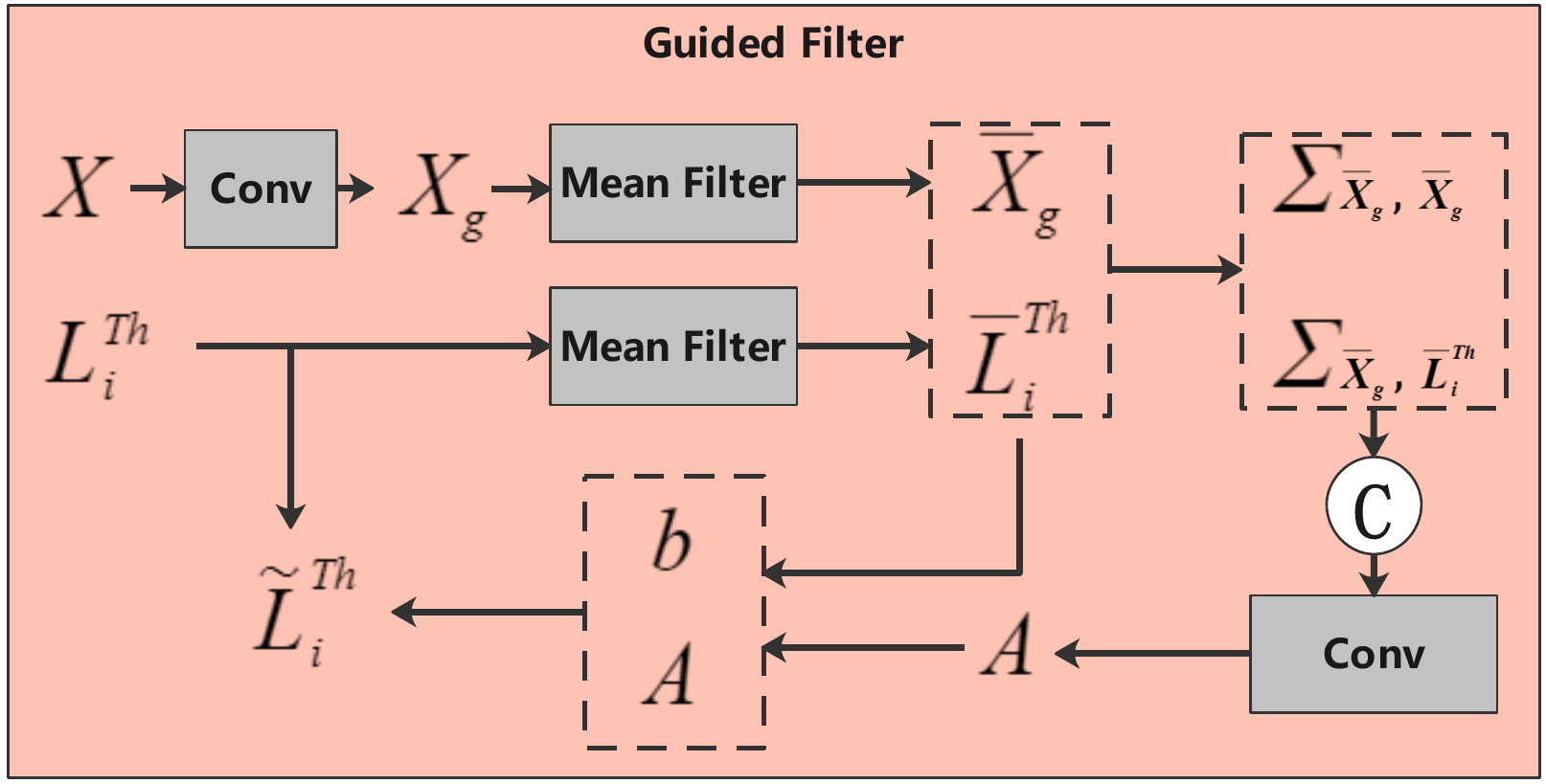}
   \end{center}
      \caption{Workflow of Guided Filter.}
   \label{fig:gf}
\end{figure}

In Figure~\ref{fig:gf}, we illustrate the workflow of the Guided Filter module.
First, a $3\times3$ convolution layer is used to map the image feature $X$ to tensor $X_g$ of three channels.
Following DGF~\cite{wu2018fast}, each instance layout $L_i^{Th}$ and $X_g$ are filtered by a prescribed $3\times3$ convolution kernel.
And the linear transformation parameters $A\in\mathbb{R}^{H\times W\times 1}$ and $b\in\mathbb{R}^{H\times W\times 1}$ are predicted from CNN layers.
Specifically, mean filter and covariance operations are carried out sequencially to get $\overline{X}_g$, $\overline{L}_{i}^{Th}$, $\Sigma_{\overline{X}_g,\overline{X}_g}$ and $\Sigma_{\overline{X}_g,\overline{L_i}^{Th}}$.
Then the parameter $A$ is predicted from $\Sigma_{\overline{X}_g,\overline{X}_g}$ and $\Sigma_{\overline{X}_g,\overline{L}_i^{Th}}$ by Convolution Block, which contains 3 conditional layers with $1\times1$ kernels.
And the parameter $b$ is equal to $\overline{L}_i^{Th}-A\odot \overline{X}_g$.
Finally, the refined layout is computed as follows:
\begin{equation}
   \widetilde{L}_i^{Th}=A\odot L_i^{Th}+b.
\end{equation}

\end{document}